\documentclass[10pt,journal,compsoc]{IEEEtran}
\usepackage[table]{xcolor}
\usepackage{booktabs,array,xcolor,colortbl,multirow,multicol,epsfig,graphicx,subfigure,amsmath,amsthm,amsfonts,mathrsfs,amssymb,bm,algorithm,algorithmic}
\usepackage{caption}
\usepackage{colortbl,accents}
\usepackage{subfigure}

\ifCLASSOPTIONcompsoc
  \usepackage[nocompress]{cite}
\else
  \usepackage{cite}
\fi

\ifCLASSINFOpdf
\else
\fi

\begin{document}
\title{Joint Modeling of Event Sequence and Time Series with Attentional Twin Recurrent Neural Networks}
%Synergic Recurrent Point Process Model for Time Series and Event Sequence with Descriptive Ability
%Synergic Recurrent Neural Network model for Time series and event sequence
%Modeling the Intensity of Point Process with Recurrent Neural Network
%Jointly Modeling Time Series and Event Sequence with Recurrent Neural Network
\author{Shuai~Xiao,~%\IEEEmembership{Member,~IEEE,}
        Junchi~Yan,~\IEEEmembership{Member,~IEEE,}
        Mehrdad~Farajtabar,~
        Le~Song,~
        %Stephen M.~Chu,~
        Xiaokang~Yang,~\IEEEmembership{Senior Member,~IEEE,}
        and~Hongyuan~Zha~%\IEEEmembership{Fellow,~OSA,}
\IEEEcompsocitemizethanks{\IEEEcompsocthanksitem S. Xiao and X. Yang are with the Department
of Electrical Engineering, Shanghai Jiao Tong University, China.\protect\\
E-mail: benjaminforever@sjtu.edu.cn, xkyang@sjtu.edu.cn
\IEEEcompsocthanksitem J. Yan (correspondence author) is with the Department of Electrical Engineering, Shanghai Jiao Tong University, China. He is also secondarily affiliated with IBM Research - China.\protect\\
E-mail: yanesta13@163.com
%\IEEEcompsocthanksitem J. Yan (correspondence author) is with the School of Computer Science and Software Engineering, East China Normal University, and IBM Research.\protect\\
%E-mail: jcyan@sei.ecnu.edu.cn
\IEEEcompsocthanksitem  M. Farajtabar, H. Zha and L. Song are with School of Computational Science and Engineering, College of Computing, Georgia Institute of Technology, Atlanta, Georgia, 30332, USA.\protect\\
E-mail: mehrdad@gatech.edu,\{zha,lsong\}@cc.gatech.edu
%\IEEEcompsocthanksitem S. M. Chu is with IBM T.J. Watson Research Center, Yorktown Heights, NY, 10598, USA.\protect\\
%E-mail: schu@us.ibm.com
%\IEEEcompsocthanksitem X. Yang is with Shanghai Key Laboratory of Media Processing and Transmission, Department of Electronic Engineering, Shanghai Jiao Tong University, Shanghai, 200240 China.\protect\\
%E-mail: xkyang@sjtu.edu.cn
}
}
%{Shell \MakeLowercase{\textit{et al.}}: Bare Demo of IEEEtran.cls for Computer Society Journals}
\IEEEtitleabstractindextext{%
\begin{abstract}
A variety of real-world processes (over networks) produce sequences of data whose complex temporal dynamics need to be studied. More especially, the event timestamps can carry important information about the underlying network dynamics, which otherwise are not available from the time-series evenly sampled from continuous signals. Moreover, in most complex processes, event sequences and evenly-sampled times series data can interact with each other, which renders joint modeling of those two sources of data necessary. To tackle the above problems, in this paper, we utilize the rich framework of (temporal) point processes to model event data and timely update its intensity function by the synergic twin Recurrent Neural Networks (RNNs). In the proposed architecture, the intensity function is synergistically modulated by one RNN with asynchronous events as input and another RNN with time series as input. Furthermore, to enhance the interpretability of the model, the attention mechanism for the neural point process is introduced. The whole model with event type and timestamp prediction output layers can be trained end-to-end and allows a black-box treatment for modeling the intensity. We substantiate the superiority of our model in synthetic data and three real-world benchmark datasets.
\end{abstract}
\begin{IEEEkeywords}
Recurrent Neural Networks, Temporal Point Process, Relational Mining, Interpretable Attention Models.
\end{IEEEkeywords}}
\maketitle
\IEEEpeerreviewmaketitle
\section{Introduction}\label{sec:introduction}
\IEEEPARstart{E}{vent} sequences are becoming increasingly available in a variety of applications.
Such event sequences, which are asynchronously generated with random timestamps, are ubiquitous in areas such as e-commerce, social networks, electronic health data, and equipment failures.
The event data can carry rich information not only about the event attribute (\emph{e.g.}, type, participator) but also the timestamp $\{z_i,t_i\}_{i=1}^N$ indicating when the event takes place.
A major line of research~\cite{AalenPP2008} has been devoted to studying event sequence, especially exploring the timestamp information to model the underlying dynamics of the system, whereby point process has been a powerful and elegant framework in this direction.

Being treated as a random variable when the event is stochastically generated in an asynchronous manner, the timestamp makes the event sequence of point processes fundamentally different from the time series~\cite{MontgomeryTSBook15} evenly-sampled from continuous signals because the asynchronous timestamps reflect the network dynamic while the time for time-series is deterministic..

However these time series data, when available, provide timely updates of background environment where events occur in the temporal point process, such as temperature for computing servers or blood pressure for patients. Many complex systems posses such time series data regularly recorded along with the point processes data.

While there have been many recent works on modeling continuous-time point processes~\cite{kdd2014Liangda,YuICDM15,farajtabar2015coevolve,DuKDD16,farajtabar2014shaping} and time series~\cite{box2015time,chatfield2016analysis, guralnik1999event}, most of them  treat  these  two  processes  independently  and  separately, ignoring the influence one may have on the other over time. To better understand the dynamics of point processes, there is an urgent need for joint models of the two processes, which are largely inexistent to date. There are related efforts in linking the time series and event sequence to each other. In fact, one popular way to convert a time series to an event sequence is by detecting multiple events (\emph{e.g.}, based on thresholding the stock price series~\cite{bacry2015hawkes}) from the series data. On the other hand, statistical aggregation (\emph{e.g.}, total number of counts) is often performed on each time interval with equal length to extract aligned time series data from the event sequences. However such a coarse treatment can lead to the key information loss about the actual behavior of the process, or at least in a too early stage.

%Recently there are many machine learning based models for scalable point process modeling~\cite{XiaoIJCAI16,farajtabar2015coevolve,du2015dirichlet}. We attribute the progressions in this direction in part to the advanced mathematical reformulations and optimization techniques \emph{e.g.},~\cite{LewisJNS2011,ZhouICML13,ZhouAISTATS13,farajtabar2015coevolve}, along with novel parametric forms for the conditional intensity function~\cite{shen2014modeling,ErtekinRPP2015,choi2015constructing,XuTKDE16} as carefully designed by researchers' prior knowledge to capture the characters of the dataset in their study. In addition to prediction tasks, the model is able to uncover the hidden structure of information diffusion network or influence patterns. Although those models provide opportunities for interpretability, one major limitation of the parametric forms of point processes is due to their specialized and restricted expression capability for arbitrary and complex event data which tends to be oversimplified or even infeasible for capturing the problem complexity in real applications. Moreover, it is prone to the risk of under-fitting/over-fitting due to model misspecification.

Recent progresses on modeling point process includes mathematical reformulations and optimization techniques~\cite{LewisJNS2011,ZhouICML13,ZhouAISTATS13,farajtabar2015coevolve} and novel parametric forms~\cite{shen2014modeling,ErtekinRPP2015,choi2015constructing,XuTKDE16} as carefully designed by human prior knowledge to capture the characters of the dataset in their study. One major limitation of those model is that the specified form of point process limits its capability to capture the dynamic of data. Moreover, it may suffer from misspecification, for which the model is not suitable for the data. Recent works \emph{e.g.},~\cite{ZhouICML13} start to turn to non-parametric form to fit the structure of a point process, but their method is under the Hawkes process formulation, which runs the risk of unknown model complexity and can be inappropriate for point processes that disobey the self or mutual-exciting rule assumed by the Hawkes model~\cite{HawkesBiometrika71}. In another recent work \cite{DuKDD16}, the authors proposed a semi-parametric pesudo point process model, which assumes a time-decaying influence between events and the background of intensity is constant. Besides, the event type is regarded as the mark associated with a univariate process.

In this paper, we view the conditional intensity of a point process as a nonlinear mapping from the joint embedding of time series and past event data to the predicted transient occurrence intensity of future events with different types. Such a nonlinear mapping is expected to be complex and flexible enough to model various characters of real event data for its application utility, \emph{e.g.}, failure prediction, social network analysis, and disease network mining as will be empirically studied in the experiment part of this paper. To overcome the disadvantages associated with the explicit parametric form of intensity function we bypass direct modeling of the intensity function and directly model the next event time and dimension. Neural networks are our choice to model this nonparametric mapping.

We utilize the state-of-the-art deep learning techniques to efficiently and flexibly model the intensity function of the point processes. Instead of predefining the form of point process, we turn to the synergic multi-RNNs (specifically twin-RNNs in this paper) as a natural way to encode such nonlinear and dynamic mapping, in an effort for modeling an end-to-end nonlinear intensity mapping without any prior knowledge. To further improve its interpretability, we infuse the attention model to improve its capability for both prediction and relational mining among event dimensions.
\begin{figure}[tb!]
	\centering
	\subfigure{\includegraphics[width=0.48\textwidth]{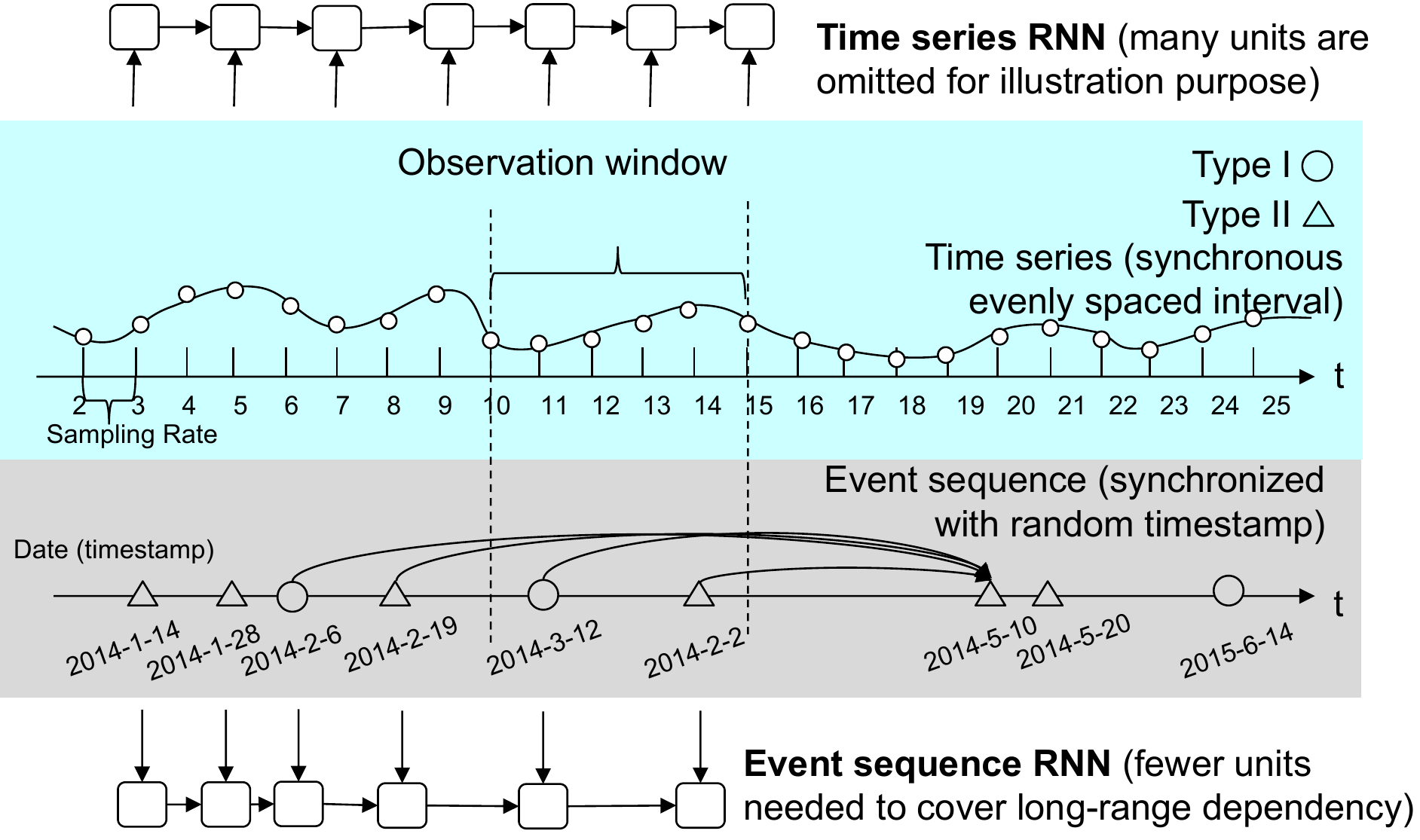}}%\vspace{-10pt}
	\caption{Time series and event sequence can be synergically modeled. The former can be used to timely capture the recent window for the time-varying features, while the latter can capture the long-range dependency over time.}
	\label{fig:ppts}
\end{figure}

\textbf{Key idea and highlights.} Our model interprets the conditional intensity function of a point process as a nonlinear mapping, which is synergetically established by a composite neural network with two RNNs as its building blocks. As illustrated in Fig.~\ref{fig:ppts}, time series (top row) and event sequence (bottom row) are distinct to each other. The underlying rationale is that time series is more suitable to carry the synchronously and regularly updated (\emph{i.e.} in a fixed pace) or constant profile features. In contrast, the event sequence can compactly catch event driven, more abrupt information, which can affect the conditional intensity function over longer period of time.

More specifically, We first argue that many conditional intensity functions can be viewed as an integration of two effects: i) spontaneous background component inherently affected by the internal (time-varying) attributes of the individual; ii) effects from history events. Meanwhile, most information in the real world can also be covered by continuously updated features like age, temperature, and asynchronous event data such as clinical records, failures. This motivates us to devise a general approach. Then, we use one RNN whose units are aligned with the time points of a time series, and another RNN whose units are aligned with events. The time series RNN can timely update the intensity function while the event sequence RNN is used to efficiently capture the long-range dependency over history. They can interact with each other through synergic non-linear mapping. This allows fitting arbitrary dynamics of point process which otherwise will be difficult or often impossible to be specified by a parameterized model restricted to certain assumptions.

As an extension to the conference version~\cite{XiaoAAAI17}\footnote{The main extensions include: i) in contrast to \cite{XiaoAAAI17} where the so-called \emph{pseudo} multi-dimensional point process~\cite{LinigerPhD2009} is adopted which involves only a single intensity function for all types of event sequence, here we separately model the intensity functions for each event type leading to the so-called \emph{genuine} multi-dimensional point process via recurrent neural networks; ii) based on the resulting multi-dimensional point process model, we incorporate a new attention mechanism to improve the interpretability of the prediction model. This expands the dependency model in RNN from the recent hidden variable $\mathbf{h}_j$ to a set of recent ones which further improves its modeling capability; iii) a more thorough analysis and verification via devised simulation experiments; iv) performing more experiments from both social network data and healthcare data. Note that the added attention model enables the relation discovery capability as also verified in both simulation based and real-world data. While the conference paper~\cite{XiaoAAAI17} only deals with event prediction rather than relation mining.}, the overall highlights of this paper are:

i) To the best of our knowledge, this is the first work to \emph{jointly} interpret and instantiate the conditional intensity function with fused \emph{time series} and \emph{event sequence} RNNs. This opens up the room for connecting the neural network techniques to traditional point process that emphasizes more on specific model driven by domain knowledge. The introduction of a full RNN treatment lessen the efforts for the design of (semi-)parametric point process model and its complex learning algorithms which often call for special tricks \emph{e.g.}~\cite{YanIJCAI13} that prohibiting the wide use for practitioners. In contrast, neural networks and specifically RNNs, are becoming off-the-shelf tools and getting widely used recently.

ii) We model the \emph{genuine} multi-dimensional point process through recurrent neural networks. Previous work~\cite{DuKDD16} use the RNN to model so-called \emph{pseudo} multi-dimensional point process~\cite{LinigerPhD2009}. Actually, they regard the event sequence as a univariate point process and treat the dimension as the mark associated with events. Consequently, there exists only one intensity function for all the processes instead of one per each dimension. On the contrary, in our work the process is the result of the superposition of sub-processes for each dimension. In this way, we can separate the parameters of each dimension as well as capture their interactions. This leads to more effective simulation and learning algorithm.

iii) To improve the interpretability, it is also perhaps the first time, to our knowledge, an attention based RNN model for point process is proposed. For multi-dimensional point process, our proposed attention mechanism allows each dimensional has its own attention parameters. One typical resulting utility involves decision support and causality analysis~\cite{XuICML16}.

iv) Our model is simple and general and can be end-to-end trained. We target three empirical application domains to demonstrate the superiority of the proposed method, namely, predictive maintenance, social network analysis and disease relation mining. The state-of-the-art performance on relational mining, event type and timestamp prediction corroborates its suitability to real-world applications.

The organization of this paper is as follows: related work is reviewed in Section \ref{sec:related}. Section \ref{sec:method} presents the main approach, and Section \ref{sec:experiment} describes the empirical studies involving three different application scenarios related to both prediction and mining. Section \ref{sec:conclusion} concludes this paper.

\section{Related Work and Motivation}\label{sec:related}
We review the related concepts and work in this section, which is mainly focused on Recurrent Neural Networks (RNNs) and their applications in time series and sequence data, respectively. Then we discuss existing point process methods and their connection to RNNs. All these observations motivate the work of this paper.

\textbf{Recurrent neural network.} The building block of our model is the Recurrent Neural Networks (RNNs)~\cite{ElmanCS90,PascanuICML13} and its modern variants \emph{e.g.}, Long Short-Term Memory (LSTM) units~\cite{HochreiterNC97,GravesArxiv13} and Gated Recurrent Units (GRU)~\cite{ChungArxiv14}. RNNs are dynamical systems whose next state and output depend on the present network state and input, which are more general models than the feed-forward networks. RNNs have long been explored in perceptual applications for many decades, however it can be very difficult for training RNNs to learn long-range dynamics in part due to the vanishing and exploding gradients problem. LSTMs provide a solution by incorporating memory units that allow the network to learn when to forget previous hidden states and when to update hidden states given new information. Recently, RNNs and LSTMs have been successfully applied in large-scale vision~\cite{GregorDraw15}, speech~\cite{GravesICML14} and language~\cite{SutskeverNIPS14} problems.

\textbf{RNNs for series and event data.} From application perspective, we consider two main scenarios in this paper: i) RNNs for synchronized series with evenly spaced interval \emph{e.g.}, time series or indexed sequence with pure order information \emph{e.g.}, language; ii) asynchronous sequence with timestamp \emph{e.g.}, event data.

\emph{i) Synchronized series}: RNNs have been a long time a natural tool for standard time series modeling and prediction~\cite{ConnorTNN94,ChandraNC12}, whereby the indexed series data point is fed as input to an (unfold) RNN. In a broader sense, video frames can also be treated as time series and RNN are widely used in recent visual analytics works~\cite{JainICRA16} and so for speech~\cite{GravesICML14}. RNNs are also intensively adopted for sequence modeling tasks~\cite{ChungArxiv14} when only order information is considered.

\emph{ii) Asynchronous event}: In contrast, event sequence with timestamp about their occurrence, which are asynchronously and randomly distributed over the continuous time space, is another typical input type for RNNs~\cite{DuKDD16,choi2016retain} (despite its title for 'time series'). One key differentiation against the first scenario is that the timestamp or time duration between events (together with other features) is taken as input to the RNNs. By doing so, (long-range) event dependency can be effectively encoded.

\textbf{Interpretability and attention model.} Prediction accuracy and model interpretability are two goals of many successful predictive methods. Existing works often have to suffer the tradeoff between the two by either picking complex black box models such as deep neural network or relying on traditional models with better interpretation such as Logistic Regression often with less accuracy compared with state-of-the-art deep neural network models. Despite the promising gain in accuracy, RNNs are relatively difficult to interpret. There have been several attempts to interpret RNNs~\cite{choi2016retain,XuICML15,cho2015describing}. However, they either compute the attention score by the same function regardless of the affected point's dimension~\cite{choi2016retain}, or only consider the hidden state of the decoder for sequence prediction~\cite{XuICML15,cho2015describing}. As for multi-dimensional point process, past events shall influence the intensity function differently for each dimension. As a result, we explicitly assign different attention function for each dimension which is modeled by respective intensity functions, thus leading to an infectivity matrix based attention mechanism which will be detailed later in this paper.

%however, they either consider binary output or the output is a dependent sequence. Here, we interpret the interaction relationship in high dimensional space for event sequences.

%Attention based neural network models have recently gained much attraction in image processing~\cite{XuICML15}, natural language processing~\cite{BahdanauICLR15} and speech recognition~\cite{chorowski2015attention}. The need for attention mechanism can be seen in the language translation task~\cite{BahdanauICLR15}: Representing the entire sentence with one fixed-size vector is inefficient; the neural translation machine usually finds it difficult to translate the given sentence represented by a single vector. In this paper, we will adopt the attention mechanism to enable the interpretability of our proposed RNN model for point process modeling.

\textbf{Point processes.} Point process is a mathematically rich and principled framework for modeling event data~\cite{AalenPP2008}. It is a random process whose realization consists of a list of discrete events localized in time. The dynamics of the point process can be well captured by its conditional intensity function whose definition is briefly reviewed here: for a short time window $[t,t+dt)$, $\lambda(t)$ represents the rate for the occurrence of a new event conditioned on the history $\mathcal{H}_t = \{z_i,t_i|t_i < t\}$:
\begin{equation}\notag
\lambda(t)=\lim_{\Delta t\rightarrow 0}\frac{\mathbb{E}(N(t+\Delta t)-N(t)|\mathcal{H}_t)}{\Delta t}=\frac{\mathbb{E}(dN(t)|\mathcal{H}_t)}{dt},
\end{equation}
where $\mathbb{E}(dN(t)|\mathcal{H}_t)$ is the expectation of the number of events happened in the interval $(t, t + dt]$ given the historical observations $\mathcal{H}_t$. The conditional intensity function has played a central role in point processes and many popular processes vary on how it is parameterized. Some typical examples include:

1) \emph{Poisson process}~\cite{KingmanPP92}: the homogeneous Poisson process has a very simple form for its intensity function: $\lambda_{d}(t)=\lambda_{d}$. Poisson process and its time-varying generalization are both assumed to be independent of the history.

2) \emph{Reinforced Poisson processes}~\cite{PemantlePS07,shen2014modeling}: the model captures the `rich-get-richer' mechanism characterized by a compact intensity function, which is recently used for popularity prediction~\cite{shen2014modeling}.

3) \emph{Hawkes process}~\cite{HawkesBiometrika71}: Recently, Hawkes process has received a wide attention in network cascades modeling~ \cite{ZhouAISTATS13,farajtabar2015coevolve}, community structure~\cite{tran2015netcodec}, viral diffusion and activity shaping\cite{farajtabar2014shaping}, criminology~\cite{lewis2010self}, optimization and intervention in social networks~\cite{farajtabar2016multistage}, recommendation systems~\cite{hosseini2017recurrent}, and verification of crowd generated data~\cite{tabibian2016distilling}. As an illustration example intensively used in this paper, we particularly write out its intensity function is:
\begin{align}\notag
\lambda_{d} &= \mu_{d}(t) + \sum_{i:t_{i}<t}\gamma_{d_{i}d}(t-t_i)  \\\notag
&= \mu_{d}(t) + \sum_{i:t_{i}<t}a_{d_{i}d}\text{exp}(-w(t-t_i)),
\end{align}
where $\mathbf{A}_{d_{i}d}=\{a_{d_{i}d}\}$ is the infectivity matrix, indicating the directional influence strength from dimension $d_i$ to $d$. It explicitly uses a triggering term to model the excitation effect from history events where the parameter $w$ denotes the decaying bandwidth. The model is originally motivated to analyze the earthquake and its aftershocks\cite{OgataJASA88}.

4) \emph{Reactive point process}~\cite{ErtekinRPP2015}: it can be regarded as a generalization of the Hawkes process by adding a self-inhibiting term to account for the inhibiting effects from history events.

5) \emph{Self-correcting process}~\cite{IshamSPP79}: its background part increases steadily, while it is decreased by a constant $e^{-\alpha}<1$ every time a new event appears.

We summarize the above forms in Table~\ref{tab:intensity}. It tries to separate the spontaneous background component and history event effect explicitly. This also motivates us to design an RNN model that can flexibly model various point process forms without model specification.
\begin{table}[tb!]
\centering
\caption{Conditional intensity functions of point processes.}
\resizebox{0.48\textwidth}{!}{
\begin{tabular}{lrr}
  \addlinespace
  \toprule
  Model&Background&History event effect\\
  \midrule
  Poisson process&$\mu(t)$&$0$\\
  Reinforced poisson process&$0$&$\gamma(t)\sum_{t_i<t}\delta(t_i<t)$\\
  %Hawkes process&$\lambda_0$&$\sum_{t_i<t}\alpha \omega\exp(-\omega(t-t_i)),\omega>0$\\
  Hawkes process&$\mu(t)$&$\sum_{t_i<t}\gamma(t,t_i)$\\
  Reactive point process&$\mu(t)$&$\sum_{t_i<t}\gamma_1(t,t_i)-\sum_{t_i<t}\gamma_2(t,t_i)$\\
  Self-correcting process&$0$&$\exp(\mu t-\sum_{t_i<t}\gamma(t,t_i))$\\
  \bottomrule
  \vspace{0.5mm}
\end{tabular}}
\small{Note:$\delta(t)$ is Dirac function, $\gamma(t,t_i)$ is the time-decaying kernel and $\mu(t)$ can be constant or time-varying function.}
\label{tab:intensity}
\end{table}

%%%%%%%%%%%%%%%%%%%%%%%%%%%%%%%%%%%%%%%%%%%%%%%%%%%%%%%%%%%%%%%%%%%%
%%%%%%%%%%%%%%%%%%%%%%%%%%%%%%%%%%%%%%%%%%%%%%%%%%%%%%%%%%%%%%%%%%%%
%%%%%%%%                    SECTION                         %%%%%%%%

\section{Network Structure and Learning}
\label{sec:method}
In this section, we will present the proposed network structure along with the learning algorithm for modeling the behavior of dynamic events.

%%%%%%%%%%%%%%%%%%%%%%%%%%%%%%%%%%%%%%%%%%%%%%%%%%%%%%%%%%%%%%%%%%%%

\subsection{Brief on RNN as building block}\label{subsec:brief}
Taking a sequence $\{\mathbf{x}\}_{t=1}^T$ as input, the RNN generates the hidden states $\{\mathbf{h}\}_{t=1}^T$, also known high-level representation of inputs~\cite{ElmanCS90,PascanuICML13}:
\begin{align}\notag
\mathbf{h}_t&= f(\mathbf{W}\mathbf{x}_t+\mathbf{H}\mathbf{h}_{t-1}+\mathbf{b}),
%\mathbf{x}^{out}_t&= \text{softmax}(\mathbf{W}\mathbf{h}_t+\mathbf{b}_y)
\end{align}%\label{eq-rnn}
where $\mathbf{x}_t$ is the profile associated with each event, and $f$ is a non-linear function, and $\mathbf{W},\mathbf{H}, \mathbf{b}$ are parameters to be learned. One common choice for non-linear function $f$ is \emph{Sigmoid} or \emph{tanh}, who suffers from vanishing-gradients problem~\cite{pascanu2013difficulty} and poor long-range dependency modeling capability. In contrast, we implement our RNN with Long Short Term Memory (LSTM)~\cite{HochreiterNC97,GravesArxiv13} for its popularity and capability for capturing long-range dependency. In fact, other RNN variants \emph{e.g.} Gated Recurrent Units (GRU)~\cite{ChungArxiv14} can also be alternative choices, while the analysis of the consequence of this particular choice is not the focus of our paper. To make the presented paper self-contained, we reiterate the formulation of LSTM as follows:
\begin{align}\notag
\mathbf{i}_t&= \sigma(\mathbf{W}_i\mathbf{x}_t+\mathbf{U}_i\mathbf{h}_{t-1}+\mathbf{V}_i\mathbf{c}_{t-1}+\mathbf{b}_{i}),\\\notag
\mathbf{f}_t&= \sigma(\mathbf{W}_f\mathbf{x}_t+\mathbf{U}_f\mathbf{h}_{t-1}+\mathbf{V}_f\mathbf{c}_{t-1}+\mathbf{b}_{f}),\\\notag
\mathbf{c}_t&= \mathbf{f}_t\mathbf{c}_{t-1}+\mathbf{i}_t\odot\text{tanh}(\mathbf{W}_c\mathbf{x}_t+\mathbf{U}_c\mathbf{h}_{t-1}+\mathbf{b}_c),\\\notag
\mathbf{o}_t&= \sigma(\mathbf{W}_o\mathbf{x}_t+\mathbf{U}_o\mathbf{h}_{t-1}+\mathbf{V}_o\mathbf{c}_{t}+\mathbf{b}_{o}),\\\notag
\mathbf{h}_t&=\mathbf{o}_t\odot\text{tanh}(\mathbf{c}_t),
\end{align}%\label{eq-lstm}
where $\odot$ denotes element-wise multiplication and the recurrent activation $\sigma$ is the Logistic Sigmoid function. Unlike standard RNNs, the Long Short Term Memory (LSTM) architecture uses memory cells to store and output information, allowing it to better discover long-range temporal relationships. Specifically, $\mathbf{i}$, $\mathbf{f}$, $\mathbf{c}$, and $\mathbf{c}$ are respectively the input gate, forget gate, output gate, and cell activation vectors\footnote{In subsection \ref{subsec:brief} we slightly abuse the notations and in fact their effects are only valid in this subsection and have no relation to other notations used in the other parts of this paper.}. By default, the value stored in the LSTM cell $\mathbf{c}$ is maintained unless it is added to by the input gate $\mathbf{i}$ or diminished by the forget gate $\mathbf{f}$. The output gate $\mathbf{o}$ controls the emission of the memory from the LSTM cell.
Compactly, we represent the LSTM system via the following equation:
\begin{equation}\notag
\mathbf{h}_t= \text{LSTM}(\mathbf{x}_t,\mathbf{h}_{t-1})
\end{equation}%\label{eq-rnn}

\begin{figure}[tb!]
	\centering
	\subfigure{\includegraphics[width=0.35\textwidth]{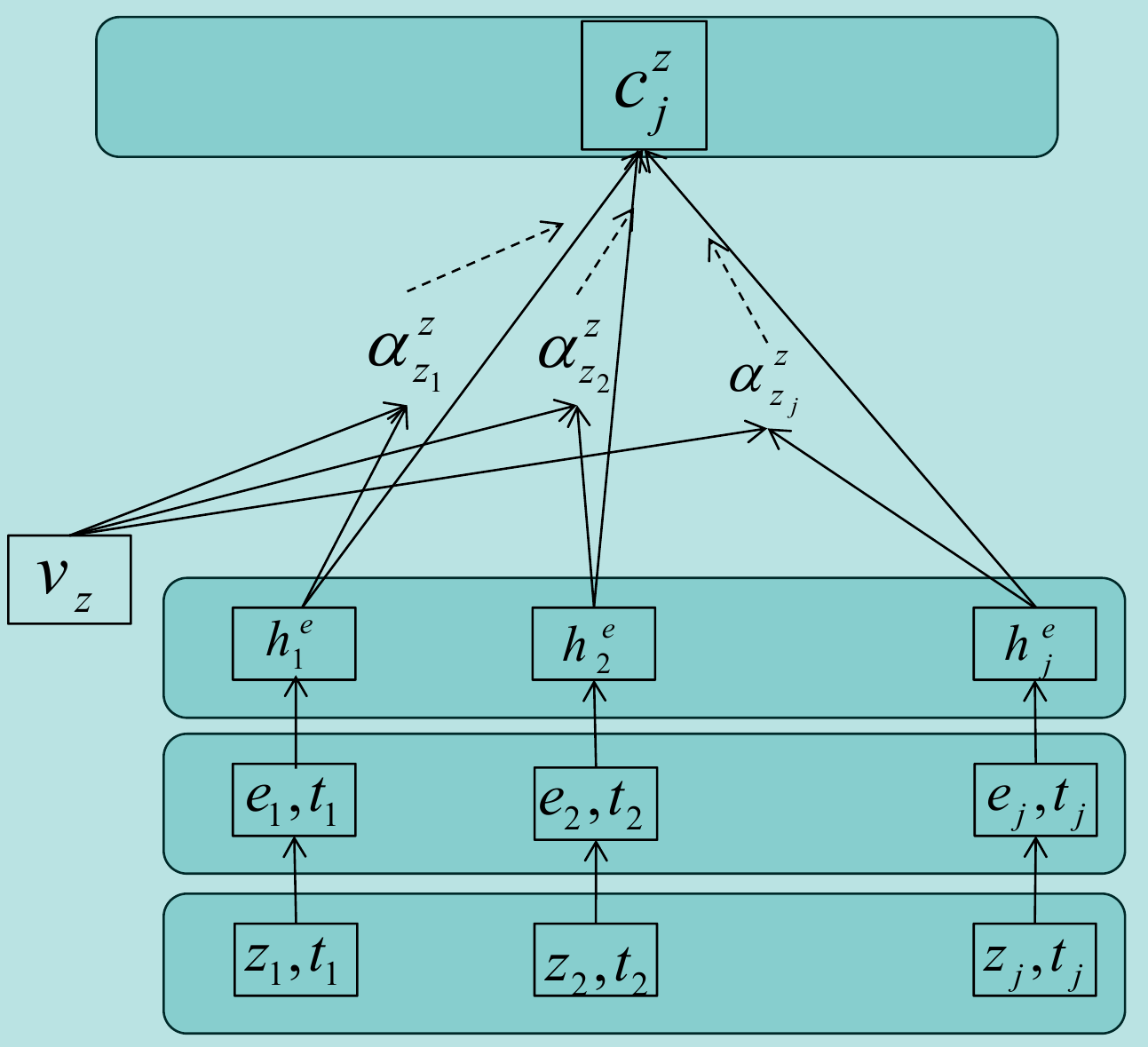}}%\vspace{-10pt}
	\caption{For a sequence of events from $1$ to $j$ (in a recent time window), $\mathbf{v}_z$ is the learned feature vector for dimension $z$, $\alpha_{z_i}^z$ is the influence strength from dimension $z_i$ to $z$, and $c_j^z$ is the new representation vector of $\mathcal{H}_{t_j}$ for $z$.}
	\label{fig:attention}
\end{figure}
\begin{figure}[tb!]
	\centering
	\subfigure{\includegraphics[width=0.48\textwidth]{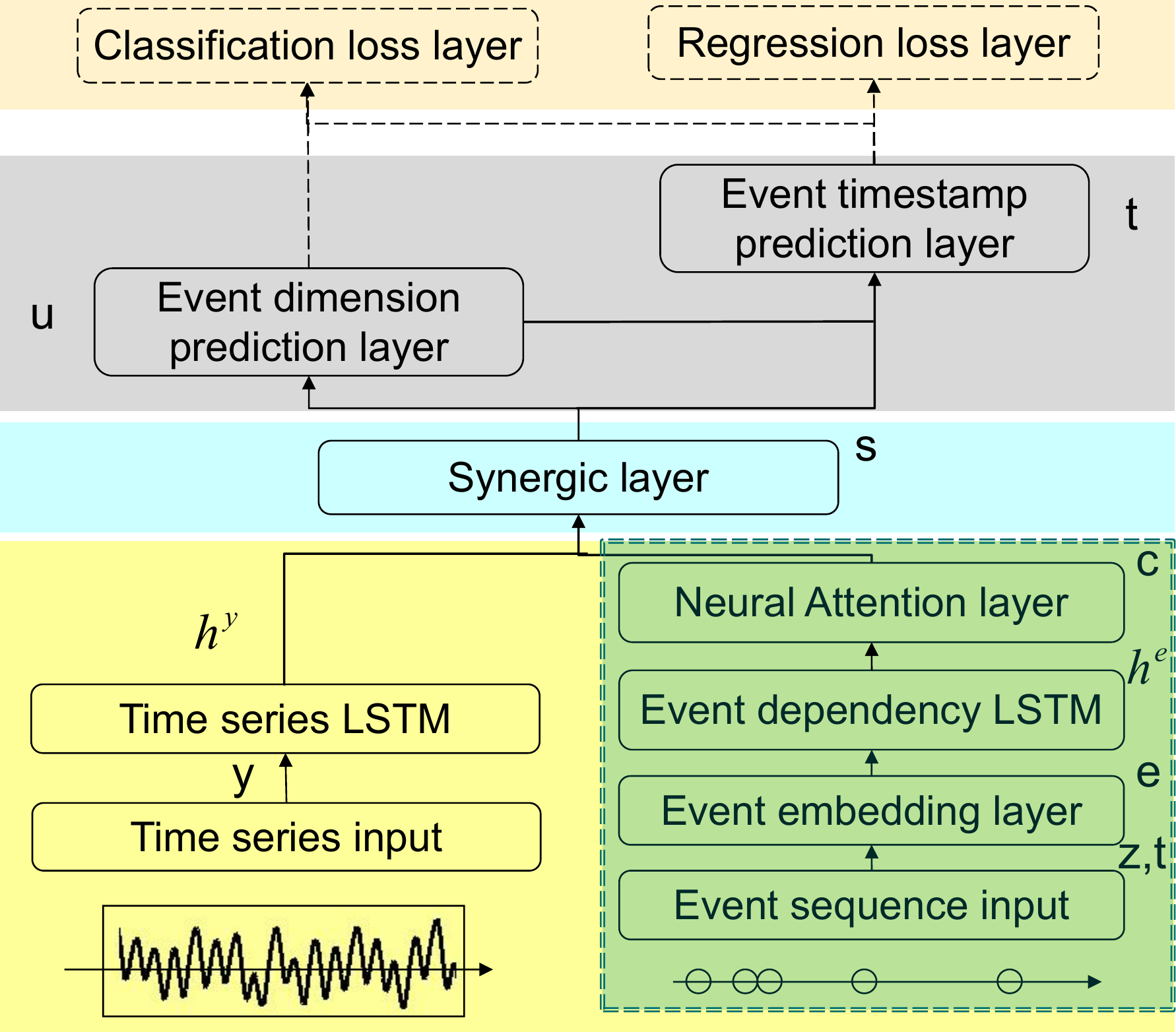}}%\vspace{-10pt}
	\caption{Our network can be trained end-to-end. Event is embedded in low-dimensional space and then pass through attention module. Time series and event sequence are connected to an synergic mapping layer that fuses the information from two LSTMs. Then output layer consists of dimension and timestamp.}
	\label{fig:overview}
\end{figure}

\subsection{Infectivity matrix based attention mechanism}
%As discussed earlier in this paper, the input data can be categorized into two types: i) time series data who are usually recorded at evenly-spaced timestamps; ii) event data whose occurrence can be arbitrary in the continuous time space. In a concrete example for analyzing patients' health records, time series can consist of temperature and blood pressure; and event data can be heart attacks and kidney failures.

Now we give further formal notational definitions used in this paper. Time series data is denoted by $\{\mathbf{y}_t\}_{t=1}^T$, \emph{e.g.} a patient's temperature and blood pressure recorded in different dimensions of the vector $\mathbf{y}_t$. Event data is represented as $\{z_i,t_i\}_{i=1}^N$, where $z_i\in \mathbb{Z}$ is the dimension representing a categorical information (\emph{e.g} a mark or type or an agent), where $\mathbb{Z}$ is the finite set of all event types, and $t_i$ is occurrence time. The former can timely affect the transient occurrence intensity of events and the latter can often abruptly cause jump and transition of states of agents and capture long-range event dependency~\cite{HawkesJAP74}.

As shown in Fig.~\ref{fig:overview}, for our proposed network, these two sources of data are fed separately to two RNNs (we call it a twin RNN structure in this paper) whose outputs are further combined to serve as the inputs of subsequent layers. For event sequence $\{z_i,t_i\}_{i=1}^N$ with length $N$, we can generate a hidden variable sequence $\{\mathbf{h}_i\}_{i=1}^N$ as the high-level representation of input sequence in RNN.
To predict the dimension (\emph{e.g.} event type) $z_{j+1}$ and time $t_{j+1}$ for the $(j+1)$-th event, the history  $\mathcal{H}_{t_j}= \{z_i,t_i\}_{i=1}^j$, prior to that event should be utilized. The most recent $\mathbf{h}_j$ is often regarded as a compressed representation of $\mathcal{H}_{t_j}$.

One may argue the necessity for involving a twin RNN structure, since the instantaneous time series data can be sampled and fed into a single RNN along with the event data when an event occurs. However, there are particular advantages for adopting such a twin-RNN structure. Empirically, it has been shown in some recent study~\cite{JainICRA16} that using separate RNN for each time series data \emph{e.g.}, video and time series sensors can lead to better prediction accuracy than a single RNN fed with the combination of the multiple time series data. More importantly, the events can occur in arbitrary timestamp \emph{i.e.}, they are asynchronous while the time series data is often sampled with equal time interval being a synchronous sequence. This inherent difference inspires us to model these two different sequence data via separate RNN as their dynamics can be rather varying.

However, there are still two limitations for the above approach: i) The prediction capability or model expressiveness is limited: in fact only the recently updated hidden variable $\mathbf{h}_j$ is used for prediction regardless of the length of input sequence. ii) The interpretability is limited. As we compress all information into a fixed vector $\mathbf{h}_j$, it is hard to infer which event contributes most to the prediction. For example in the problem of multi-dimensional Hawkes process learning, one important goal is to uncover the hidden network structure, infectivity matrix $A$, from real-world event sequences, such as the influence strength between users in social network~\cite{ZhouAISTATS13,farajtabar2015coevolve}, or progression relationship between event types~\cite{choi2015constructing}. Uncovering the hidden structure is also stressed in causal analysis~\cite{XuICML16}, which gives much evidence for prediction result. This calls for particular mechanisms to improve its flexibility and interpretability.

In this work, we devise a temporal attention mechanism to enable interpretable prediction models for point processes. Events from a certain dimension may have higher influence on some dimensions. We exploit this observation to make the trained neural network model more interpretable and expressive. To achieve this, we first expand the representation of $\mathcal{H}_{t_j}$ to be a set of vectors $\{\mathbf{h}_i\}_{i=1}^j$ instead of only the most recent $\mathbf{h}_j$, referred as context vectors. Each of them is localized to its respective preceding event of interest from inputs $\{z_i,t_i\}_{i=1}^j$. Inspired by the Hawkes process, the influence strength $ \alpha_{z_i}^z$ from $z_i$ to $z$ is introduced and it is modeled by:
\begin{equation}\label{eq:context_score}
\alpha_{z_i}^z = f_{att}(\mathbf{h}_i,\mathbf{v}_z)
\end{equation}
where $\mathbf{v}_z$ is the feature vector to be learned for the particular prediction dimension $z$ and $f_{att}$ is the score function which gives the influence strength from $z_i$ to $z$. Once the influence strength $\alpha_{z_i}^z$ are computed, we can generate the representation vector $\mathbf{c}_j^z$ for the next layer:
\begin{equation}
\mathbf{c}_j^z = \phi(\{\mathbf{h}_i\}_{i=1}^j,\{\alpha_{z_i}^z\}_{i=1}^j),
\end{equation}\label{eq:generate_context}
where $\phi$ is the attention function, which computes the final representation of $\mathcal{H}_{t_j}$.
Here we choose the widely used soft attention mechanism~\cite{XuICML15}, whose influence from former events is in an additive form~\cite{farajtabar2015coevolve}:
\begin{equation}\label{eq:context_vector}
\mathbf{c}_j^z = \sum_{i=1}^S \alpha_{z_i}^z \mathbf{h}_i
\end{equation}
Note that hard attention~\cite{cho2015describing} only assigns 0 or 1 to the influence strength $ \alpha_{z_i}^z$, which is too rough to capture fine-grained influence. Since the cost is differentiable with respect to the parameters, we can easily train the network end-to-end using backpropagation.

After the model is trained \emph{i.e.} the parameter $\mathbf{v}_z$ is fixed, for each testing event sequence $k$ and its computed $\{\{\alpha_{z_i}^z\}_{i=1}^j\}_k$ by Eq.\ref{eq:context_score}, we define the infectivity matrix to reflect the mutual influence among dimensions as $\mathbf{A}_{z_i,z}=\langle \alpha_{z_i}^{z} \rangle, z_i,z\in \mathbb{Z}$, where $\langle \cdot \rangle$ represents the average of all $\{\alpha_{z_i}^{z}\}_k$ divided by $k$.

The attention mechanism is depicted in Fig.~\ref{fig:attention}. This attention mechanism can allow the network to refer back to the preceding hidden variables $\{\mathbf{h}_j\}_{i=1}^j$, instead of forcing it to encode all information into the recent $\mathbf{h}_j$. It can retrieve from internal memory and choose what to attend to.

Finally we point out that in the context of point process, our attention mechanism is different from the existing work~\cite{choi2016retain,XuICML15,cho2015describing} in that they only consider one-way effect over the sequence. In another word, their approaches are current state agnostic. This leads to a vector representation (similar to the role of the vector $\mathbf{v}_z$ used in Eq.\ref{eq:context_score}) for the weight variables $\mathbf{\alpha}$ instead of a two-way infectivity matrix $\mathbf{A}$. Moreover, to make the model tractable, we use a parameterized form by Eq.\ref{eq:context_score} to represent the two-way weights.

\subsection{Network structure}
Now we give the full description of our network whose overview is depicted in Fig.~\ref{fig:overview}. The dashed part in the right of figure is illustrated in detail in Fig.~\ref{fig:attention}. For time series data \emph{e.g.}, temperature, blood pressure, they are sampled evenly over time. We use $\{\mathbf{y}_t\}_{t=1}^T$ to indicate the dense feature vector sampled at different timestamps. Those signals are expected to reflect the states of each dimension and drive the occurrence intensity of events. Hence we have:
\begin{equation}
\label{eq:continuous_lstm}
\notag
\mathbf{h}^y_t= \text{LSTM}_{y}(\mathbf{y}_t,\mathbf{h}^y_{t-1})
\end{equation}
For event sequence $\{z_i,t_i\}_{i=1}^N$, we can generate hidden states through LSTM, which can capture long-range dependency of events. First we project the dimension $z_i$ to a low-dimensional embedding vector space. Then the embedding vector combined with timestamps is fed to LSTM. We use the following equation to represent the process:
\begin{align}\notag
\label{eq:event_lstm}
&\mathbf{e}_i = \mathbf{W}_{em}\mathbf{z}_i,\\
&\mathbf{h}_i^e = \text{LSTM}_{z}(\{\mathbf{e}_i,\mathbf{t}_i\},\mathbf{h}_{i-1}^e), \notag
\end{align}
where $\mathbf{e}_i$ denotes the embedding vector of input $\mathbf{z}_i$ and $\mathbf{W}_{em}$ the embedding matrix to learn.

For dimension $z$, its final representation of $\mathcal{H}_{t_j}$ is $\mathbf{c}_j^z$, which is obtained through the attention mechanism introduced in Eq.~\ref{eq:context_vector}.
For the score function of Eq.~\ref{eq:context_score}, we specify it by:
\begin{equation}
\label{eq:score_fun}
f_{att}(\mathbf{h}_i^e,\mathbf{v}_z)=
\begin{cases}
  0, & \text{if}\ |\text{tanh}(\mathbf{h}_i^e*\mathbf{v}_z)| < \epsilon \\
  |\text{tanh}(\mathbf{h}^e_i*\mathbf{v}_z)|, & \text{otherwise}
\end{cases}
\end{equation}

When the context vector $\mathbf{h}_i^e$ is similar to $\mathbf{v}_z$, the attention function $f_{att}$ produces a large score. Otherwise a small one. In an extreme case, the score is zero when the context vector is orthogonal to feature vector $\mathbf{v}_z$ of dimension $z$. To promote sparsity of infectivity matrix, we threshold the score with a minus $\epsilon$ operation. The threshold $\epsilon$ can control the degree of sparsity which is set to $0.01$ throughout this paper. Note the form of Eq.~\ref{eq:score_fun} is also used in~\cite{XuICML15,cho2015describing} to model the one-way attention weight $\alpha_i=f_{att}(\mathbf{h}_i,\mathbf{v})$. From this, it is clear that our model for the attention is two-way between dimension $i$ to $z$ as shown in Eq.~\ref{eq:context_score}.

To jointly model event sequence and time series, we combine them into one synergic layer as illustrated in Fig.~\ref{fig:overview}:
\begin{equation}
\mathbf{s}^z_{j} = \text{f}_{syn}(\mathbf{W}_f[\mathbf{h}^y_{t_j},\mathbf{c}_j^z]+\mathbf{b}_f),
\end{equation}%\label{eq-lstm}
where $[\mathbf{h}^y_{t_j},\mathbf{c}_j^d]$ is the concatenation of the two vectors. The synergic layer can be any function, coupling two data sources together. Here we use the Sigmoid function. As a result, we obtain a representation $\mathbf{s}^z_j$ for the output dimension $z$. We can use this representation to compute the intensity for each dimension and then simulate its next occurrence time and its dimension jointly. Here we take a more efficient approach by firstly predicting the next event's dimension and then further predicting the occurrence timestamp based on the predicted event dimension. Note that the intensity function is modeled implicitly within the neural network architecture and we directly model the timing and dimension of the events. In this way, we overcome the expensive computation cost from explicit parametric form of intensity function.

To predict the next event's dimension $\mathbf{u}_{j+1}$, we apply the Softmax operation to those representations $\{\mathbf{s}^z_j\}_{z=1}^{Z}$ where $Z$ is the number of event dimensions.
\begin{equation}
\mathbf{u}_{j+1} =\text{softMax}(\mathbf{w}_u\mathbf{s}_j^1,\ldots,\mathbf{w}_u\mathbf{s}_j^Z)
\end{equation}
where $\mathbf{w}_u$ are model parameters to learn. The optimal dimension $z_{j+1}^*$ (as the prediction result) is computed by selecting the corresponding maximum element in $\mathbf{u}_{j+1}$:
\begin{equation}
 z_{j+1}^* =  \underset{d}{\operatorname{argmax}} \quad \mathbf{u}_{j+1}
\end{equation}

After we obtain the optimal dimension $z_{j+1}^*$, we use the representation $\mathbf{s}^{z_{j+1}^*}_{j}$ to derive occurrence time following:
\begin{equation}
 {t}_{j+1}^* = \mathbf{w}_s\mathbf{s}_j^{ z_{j+1}^*}+{b}_s,
\end{equation}
where $\mathbf{w}_s$ are model parameters for learning.
\subsection{End-to-end learning}
The likelihood of observing a sequence $\{z_i,t_i\}_{i=1}^N$ along with time series signals $\{\mathbf{y}_t\}_{t=1}^T$ can be expressed as follows:
\begin{small}
\begin{equation}
 \mathcal{L}\big(\{z_i,t_i\}_{i=1}^N\big) = \sum_{j=1}^{N-1}\{\mathbf{b}_{z_{j+1}}\log(\mathbf{u}^{z_{j+1}}_{j+1})+\log\left(f(t_{j+1}|\mathcal{H}_{t_j})\right)\}
\label{eq:loss}
\end{equation}
\end{small}
where the weight parameters $\mathbf{b}$ are set as the inverse of the sample count in that dimension against the total size of samples, to weight more on those dimensions with fewer training samples. This is in line with the importance weighting policy for skewed data in machine learning~\cite{rosenberg2012classifying}.

For the second term, the underlying rationale is that we not only encourage correct prediction of the coming event dimension, but also require the corresponding timestamp of the event to be close to the ground truth. We adopt a Gaussian penalty function:
\begin{equation}\notag
f(t_{j+1}|\mathcal{H}_{t_j})=\frac{1}{\sqrt{2\pi\sigma}}\exp\left(\frac{-(t_{j+1}-t^*_{j+1})^2}{2\sigma^2}\right)
\end{equation}

As shown in Fig.~\ref{fig:overview}, the output $t^*_{j+1}$ from the timestamp prediction layer is fed to the classification loss layer to compute the above penalty given the actual timestamp $t_{j+1}$. We adopt RMSprop gradients~\cite{RmspropArxiv15} which have been shown to work well on training deep networks to learn these parameters.

By directly optimizing the loss function, we learn the prediction model in an end-to-end manner without the need for sophisticated or carefully designed algorithms (\emph{e.g.}, Majorization-Minimization techniques~\cite{YanIJCAI2013,farajtabar2015coevolve}) used in generative Point process models. Moreover, as pointed out by recent work~\cite{XuTKDE16}, another limitation for the generative point process model is that they are aimed to maximize the joint probability of all observed events via a maximum likelihood estimator, which is not tailored to the prediction task.

%%%%%%%%%%%%%%%%%%%%%%%%%%%%%%%%%%%%%%%%%%%%%
%%%%%%%%%%%%%%%%%%%%%%%%%%%%%%%%%%%%%%%%%%%%%

\section{Experiments and Discussion}\label{sec:experiment}

We evaluate the proposed approach on both synthetic and real-world datasets, from which three popular application scenarios are covered: social network analysis, electronic health records (EHR) mining, and proactive machine maintenance. The first two scenarios involve public available benchmark dataset: MemeTracker and MIMIC, while the latter involves a private ATM maintenance dataset from a commercial bank headquartered in North America.

The code is based on Theano running on a Linux server with 32G memory, 2 CPUs with 6 cores for each: Intel(R) Xeon(R) CPU E5-2603 v3@1.60GHz. We also use 4 GPU:GeForce GTX TITAN X with 12G memory backed by CUDA and MKL for acceleration.
\subsection{Baselines and evaluation metrics}
We compare the proposed method to the following algorithms and state-of-the-art methods:

1) \textbf{Logistic model}: We use Logistic regression for event timestamp prediction and an independent Logistic classification model for event type prediction. To make sure the proposed method and the logistic model use the same amount of information, the predictor features in the regression are comprised of the concatenation of feature vectors for sub-windows of all active time series RNN.

2) \textbf{Hawkes Process}: To enable multi-type event prediction, we use a Multi-dimensional Hawkes process~\cite{ZhouAISTATS13,farajtabar2014shaping}. The Full, Sparse, LowRankSparse indicate the different types of Hawkes Process model. The full model has no regularization on infectivity matrix while the Sparse and LowRankSparse ones have sparse and lowrank-sparse regularization, respectively. In the following, Hawkes process indicates LowRankSparse model if not explicitly mentioned. The inputs are comprised of event sequences with dimensions and timestamps.

3) \textbf{Recurrent Marked Temporal Point Processes (RMTPP)}:~\cite{DuKDD16} uses a neural network to model the event dependency flexibly. The inputs are event sequences with continuous signals sampled when they happen. The method can only sample features of transient time series  when the events happen and use partially parametric form for the base intensity and a time-decaying influence from former event to the current one. Another difference is that it assumes an independent distribution between time and marks and predicts the dimension and time independently given the learned representation of the history.

4) \textbf{TRPP}: This method uses time series and event sequences to collaboratively model the intensity function of point process. The model uses RNN to model the non-linear mapping from history to the predicted marker and time, and treats RNN as a block box without much interpretability. Here we rename it as \textbf{Twin Recurrent Point Processes (TRPP)}. This method is the one presented in the conference version of this paper~\cite{XiaoAAAI17}.

5) \textbf{ERPP}: We also include \textbf{TRPP}'s degraded version by only keeping the event sequence as input and removing the time series RNN, which is termed by \textbf{Event Recurrent Point Processes (ERPP)}. Including the term `event' also helps distinguish it from the existing term RPP: Reinforced Poisson Processes~\cite{PemantlePS07,shen2014modeling}.

6) \textbf{Markov Chain (MC)}: The Markov chain refers to the sequence of random variables, with the Markov property that future state only depends on the current state and is conditionally independent of the history. The order of Markov chain indicates how many recent states on which the future state depends. As this model can only learn the transition probability of dimensions, we use it to predict the dimensions without taking the time into account. The optimal order of Markov chain is determined by the performance on separate validation dataset.

7) \textbf{Continuous Time Markov Chain (CTMC)}: The CTMC is a special type of semi-Markov model, which models the continuous transition among dimensions as a Markov process. It predicts the next dimension with the earliest transition time, therefore, it can jointly predict time and dimension.

8) \textbf{Homogeneous Poisson Process}: This method implements the most basic point process model in which the intensity function is constant and events occur independently. It can estimate interval-event gaps.

9) \textbf{Self-correcting Process}: When the occurrence of an event decrease the probability of other events, we are facing a variant of point process called self-correcting processes. Its intensity function is shown in Table~\ref{tab:intensity} and it can estimate the inter-event time.

Inline with the above TRPP and ERPP methods, we term our model \textbf{Attentional Twin Recurrent Point Processes (ATRPP)}. To further study the effect of the time series RNN, we also evaluate the baseline version without using this channel, which is termed as \textbf{Attentional Event Recurrent Point Processes (AERPP)}.

\textbf{Evaluation metrics}. We use several common metrics for performance evaluation. For the next event dimension prediction, we adopt \emph{Precision}, \emph{Recall}, \emph{F1 Score} and \emph{Confusion matrix}. For event time prediction, we use the \emph{Mean Absolute Error (MAE)} which measures the absolute difference between the predicted time point and the actual one. For the infectivity matrix, we use \emph{RankCorr}~\cite{ZhouAISTATS13} to measure whether the relative order of the estimated influence strength is correctly recovered, when the true infectivity matrix is available. The \emph{RankCorr} is defined as the averaged Kendall rank correlation coefficient\footnote{https://en.wikipedia.org/wiki/Kendall\_rank\_correlation\_coefficient} between each row of ground-truth and estimated infectivity matrix. \emph{RelErr} measures the relative deviation between estimated $a_{ij}^*$ and and ground-truth $a_{ij}$, which is defined as the average of $\frac{|a_{ij}^*-a_{ij}|}{a_{ij}}, i,j\in \mathbb{Z}$.
\begin{figure*}[tb!]
\centering
  \subfigure{\label{fig:syn_mae}
  \includegraphics[width=0.24\textwidth]{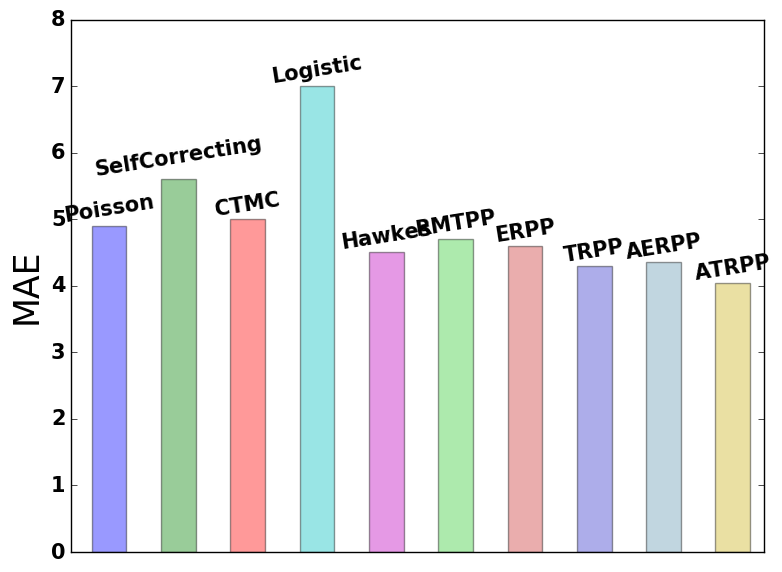}}
  \subfigure{\label{fig:syn_accracy}
  \includegraphics[width=0.24\textwidth]{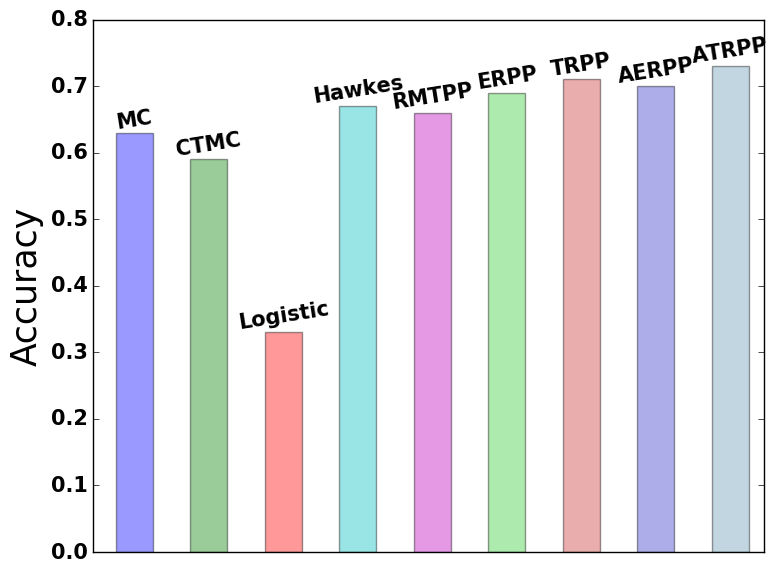}}
  \subfigure{\label{fig:syn_RelErr}
  \includegraphics[width=0.24\textwidth]{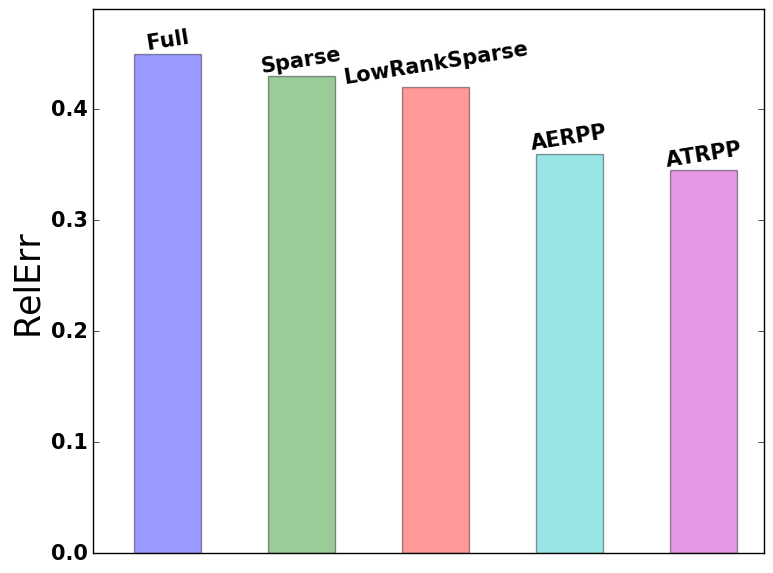}}
  \subfigure{\label{fig:syn_rankcorr}
  \includegraphics[width=0.24\textwidth]{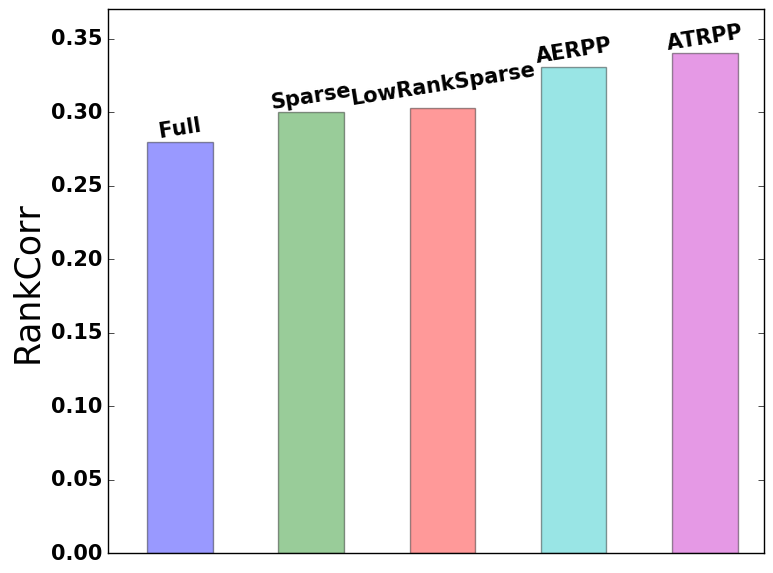}}
	\caption{Performance for MAE, Accuracy, RelErr, RankCorr over the synthetic data.}
	\label{fig:synthetic_performance}
\end{figure*}

\subsection{Experiments on synthetic data}
The test on synthetic data provides a quantitative way for evaluating the performance of compared methods.

\textbf{Synthetic data generation}. We use simulated data with known ground-truth to quantitatively verify our model by aforementioned metrics. Specifically, we generate cascades from multi-dimensional hawkes process via the Thinning algorithm~\cite{ogata1981lewis}. We choose $Z=20$ for the number of event dimensions. The background intensity term is set uniformly at random: $\mu_d \sim U(0,0.01)$. Mutual influence is set similarly to $a_{ij}\sim U(0,0.1)$. Half of the elements in the infectivity matrix are set to 0 by random in order to mimic the sparsity of influence between dimensions in many real-world problems. For the stability of the simulation process, the matrix is scaled such that its spectral radius is no larger than one. The decaying parameter in the Hawkes process is set to $w=0.01$. The detailed description of these parameters can be found in Sec.~\ref{sec:introduction} for the Hawkes process. We simulate 5000 independent cascades, 3000 for training, 1000 for validating, and 1000 for testing, respectively.
To generate time series signals, $y$, with some explanation capability to the background intensity, we sample from $y=\mu_d+n_d$ for all dimensions $d$, where, $n_d$ is a Gaussian noise, $n_d \sim U(0,0.001)$.

\textbf{Experimental results}. The performance on synthetic data is shown in Fig.~\ref{fig:synthetic_performance}. The relative error of infectivity matrix, RankCorr is demonstrated to verify the capability of uncovering hidden network structure among those dimensions \emph{i.e.}, the nodes in the network. The accuracy of time and dimension prediction is shown in order to compare the predictive performance. Our model achieves a better prediction performance, and meanwhile uncovers the infectivity matrix better than the alternatives. The self-correcting process suffers from model misspecification and performs worse than other point process models as the events are self-exciting not self-correcting. Our non-parametric model can learn from data and generalize well without prior knowledge of data.

\subsection{Predictive machine maintenance}
Predictive maintenance is a sound testbed for our model. It involves equipment risk prediction to allow for proactive scheduling of corrective maintenance. Such an early identification of potential concerns helps deploy limited resources more efficiently and cost effectively, reduce operations costs and maximize equipment uptime. Predictive maintenance is adopted in a wide variety of applications such as fire inspection, data center and electrical grid management e.g. \cite{ErtekinRPP2015}. For its practical importance in different scenarios and relative rich event data for modeling, we target our model to a real-world dataset of more than 1,000 automated teller machines (ATMs) from a global bank headquartered in North America.

We have no prior knowledge on the dynamics of the complex system and the task can involve arbitrarily working schedules and heterogeneous mix of conditions. It takes much cost or even impractical to devise specialized models.

%In maintenance support services, when a device fails, the equipment owner raises a maintenance service ticket and technician will be assigned to repair the failure. In fact, the history log and relevant profile information about the equipment can be indicative signals for the coming failures.

The studied dataset is comprised of the event logs involving error reporting and failure tickets, which is originally collected from 1,554 ATMs.
%The dataset consists of 1554 Wincor ATMs that cover 5 ATM machine models: ProCash 2100 RL (1410), 1500 RL (24), 2100 FL (74), 1500 FL (36), and 2250XE RL (10).
The event log of error records includes device identity, timestamp, message content, priority, code, and action. A ticket (TIKT) means that maintenance will be conducted. Statistics of the data is presented in Table~\ref{tab:type_statistics}. The error type indicates which component encounters an error:
1) printer (PRT), 2) cash dispenser module (CNG), 3) Internet data center (IDC), 4) communication part (COMM), 5) printer monitor (LMTP), 6) miscellaneous \emph{e.g.}, hip card module, usb (MISC). The time series here consists of features: i) the inventory information: ATM type, age, operations, temperatures; ii) event frequency for each event in the recent an hour interval. The event types and their occurrence time from the ATMs are an event sequences. Therefore, there are 1554 sequences in total, which are randomly divided into training (50\%), validating (20\%) and testing (30\%) portions.

\begin{table}[t]
\centering
\caption{Statistics of event count per ATM, and timestamp interval in days for all ATMs (in brackets).}%
\resizebox{0.48\textwidth}{!}{
\begin{tabular}{lrrrrr}
  \toprule
type&total&max&min&mean&std\\
  \midrule
  TIKT&2226(--)&10(137.04)&0(1.21)&2.09(31.70)&1.85(25.14)\\
  %Error&28434(--)&168(153.90)&0(0.10)&26.70(6.31)&18.38(9.74)\\
  PRT&9204(--)&88(210.13)&0(0.10)&8.64(12.12)&11.37(21.41)\\
  CNG&7767(--)&50(200.07)&0(0.10)&7.29(15.49)&6.59(23.87)\\
  IDC&4082(--)&116(206.61)&0(0.10)&3.83(23.85)&5.84(30.71)\\
  COMM&3371(--)&47(202.79)&0(0.10)&3.16(22.35)&3.90(29.36)\\
  LMTP&2525(--)&81(207.93)&0(0.10)&2.37(22.86)&4.41(34.56)\\
  MISC&1485(--)&32(204.41)&0(0.10)&1.39(24.27)&2.54(34.38)\\
 \bottomrule
\end{tabular}}
\label{tab:type_statistics}
\end{table}

\begin{table}[t]
\centering
\caption{Prediction performance evaluation on the ATM maintenance dataset.}
\resizebox{0.48\textwidth}{!}{
\begin{tabular}{lrrrr}
 \toprule
  model &precision &recall &F1 score & MAE\\
 \midrule
  Poisson & ----- & -----  & ----- & 4.76\\
  SelfCorrecting & ----- & -----  & ----- & 4.65 \\
  Markov Chain & 0.530 &0.591 & 0.545  & -----\\
  CTMC & 0.516&0.554& 0.503& 5.16\\
  Logistic & 0.428 & 0.375 & 0.367 & 4.51\\
  Hawkes & 0.459 & 0.514 & 0.495 & 5.43\\
  RMTPP &  0.587& 0.640 & 0.607&  4.31\\
  TRPP &  0.607&  0.661&  0.626& 4.18\\
  ERPP &  0.559& 0.639& 0.599&  4.37\\
  ATRPP &  $\mathbf{0.615}$&$\mathbf{0.688}$& $\mathbf{0.634}$& $\mathbf{3.92}$  \\
  AERPP &  0.599& 0.672& 0.617&  3.98\\
 \bottomrule
 \end{tabular}}
\label{tab:performance}
\end{table}

\begin{figure*}[t]
\centering
  \subfigure[\scriptsize ATRPP]{\label{fig:atm_isrpp}
  \includegraphics[width=0.26\textwidth]{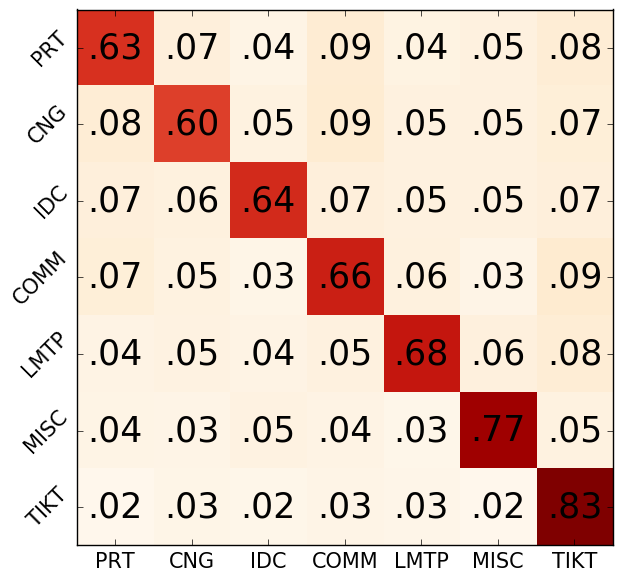}}
  \subfigure[\scriptsize AERPP]{\label{fig:srnn}
  \includegraphics[width=0.26\textwidth]{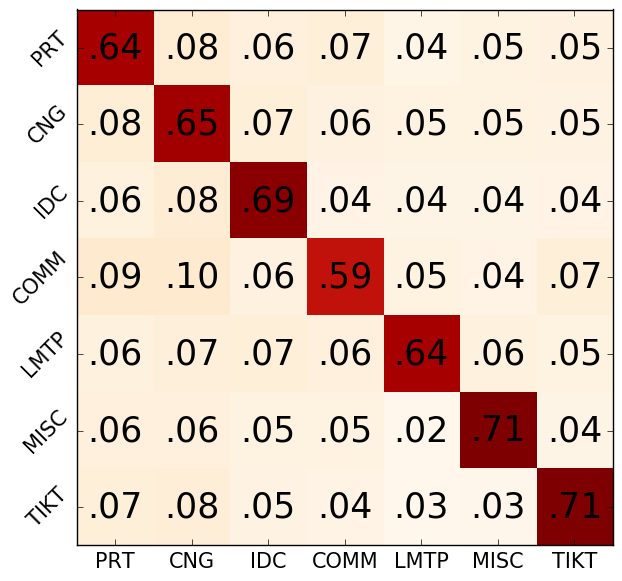}}
  \subfigure[\scriptsize Hawkes]{\label{fig:atm_hawkes}
  \includegraphics[width=0.26\textwidth]{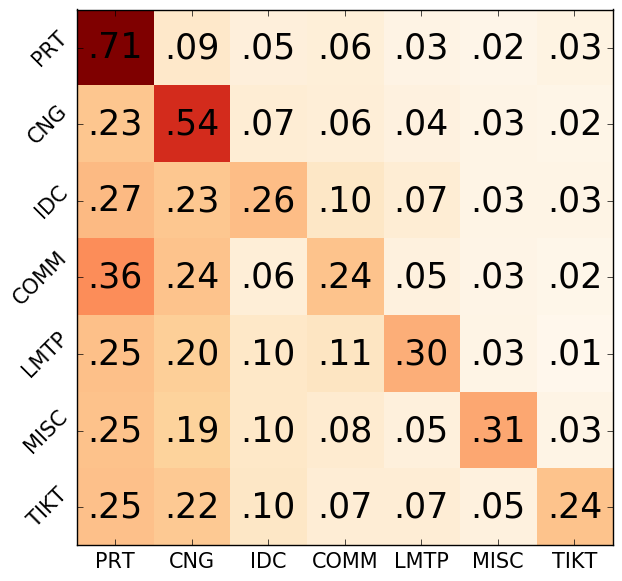}}\\
  \subfigure[\scriptsize TRPP]{\label{fig:atm_irpp}
  \includegraphics[width=0.26\textwidth]{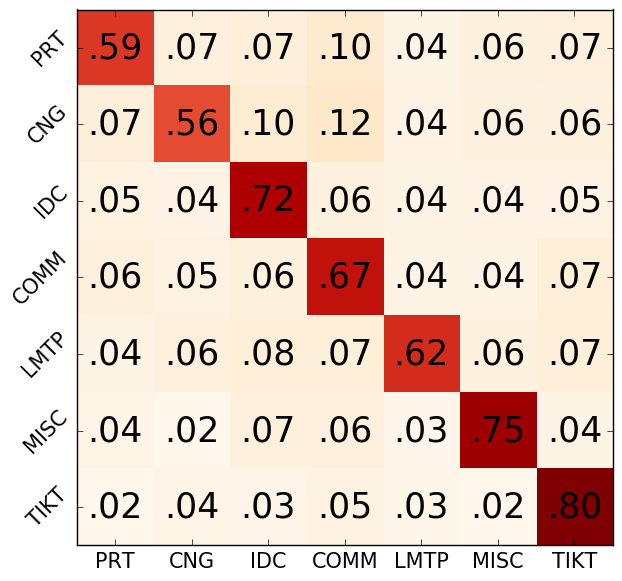}}
  \subfigure[\scriptsize ERPP]{\label{fig:atm_ernn}
  \includegraphics[width=0.26\textwidth]{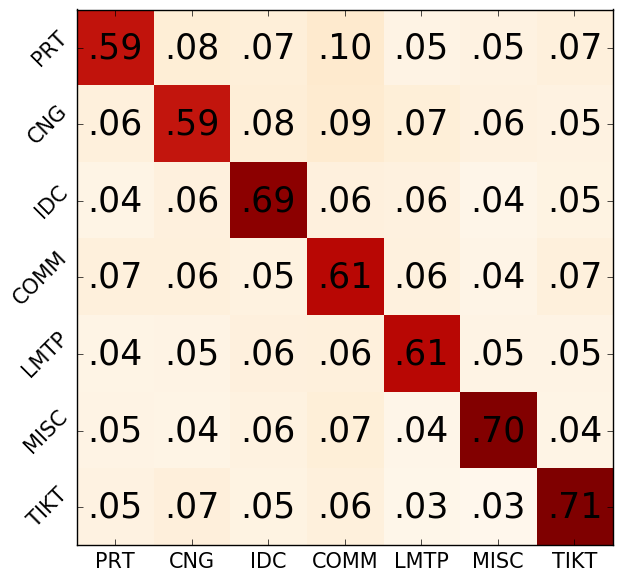}}%\hspace{-2pt}
  \subfigure[\scriptsize RMTPP]{\label{fig:atm_rmtpp}
  \includegraphics[width=0.26\textwidth]{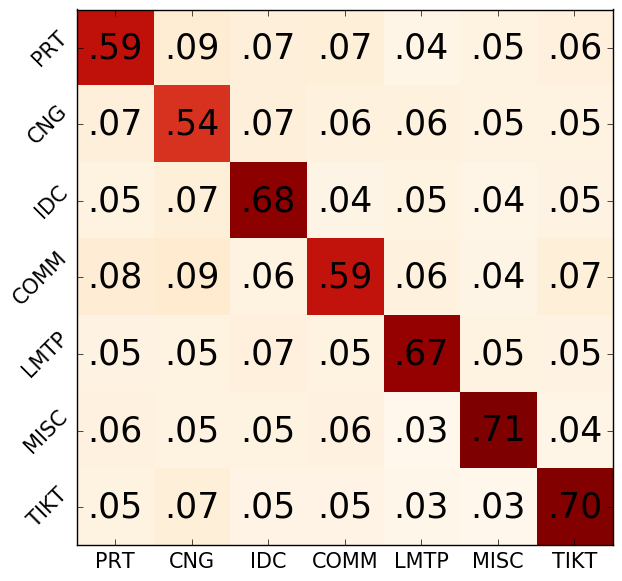}}\\%\vspace{-8pt}
  \subfigure[\scriptsize CTMC]{\label{fig:atm_cmtc}
  \includegraphics[width=0.26\textwidth]{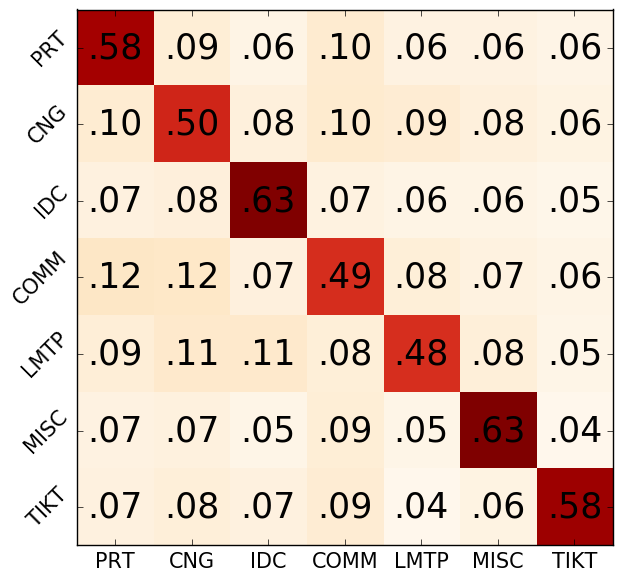}}%\hspace{-2pt}
  \subfigure[\scriptsize Markov Chain]{\label{fig:atm_mc}
  \includegraphics[width=0.26\textwidth]{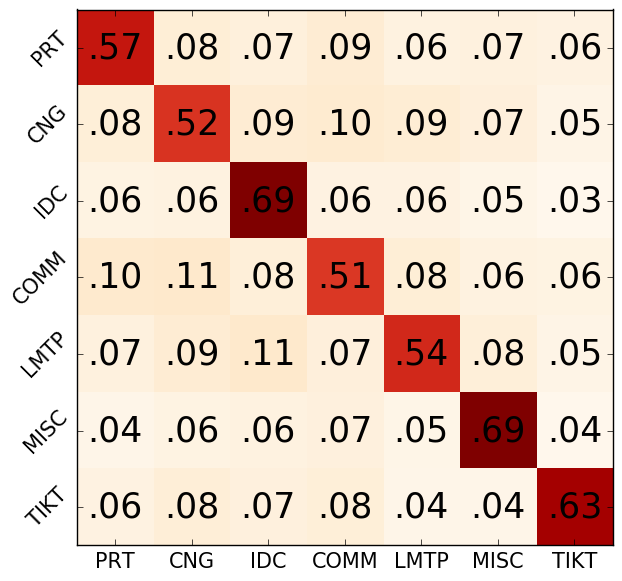}}%\hspace{-2pt}
  \subfigure[\scriptsize Logistic]{\label{fig:atm_logistic}
  \includegraphics[width=0.26\textwidth]{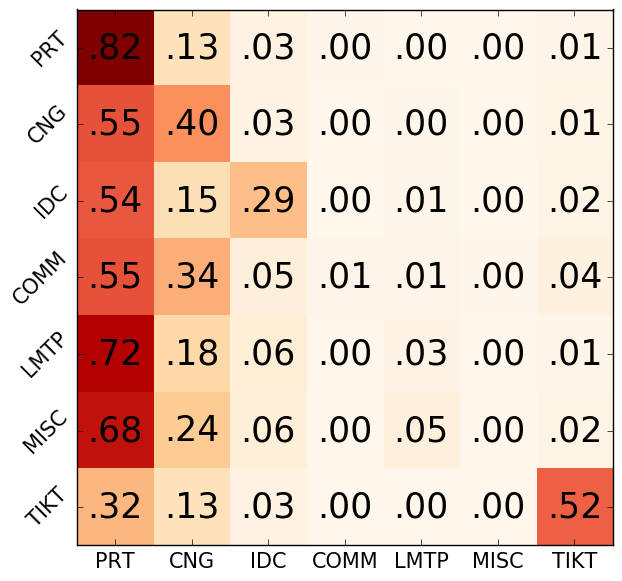}}\\%\hspace{-2pt}
	\caption{Confusion matrices over different event dimensions on the ATM maintenance dataset.}
	\label{fig:confusion}
\end{figure*}

\begin{figure}[t]
	\centering
	\subfigure{\includegraphics[width=0.32\textwidth]{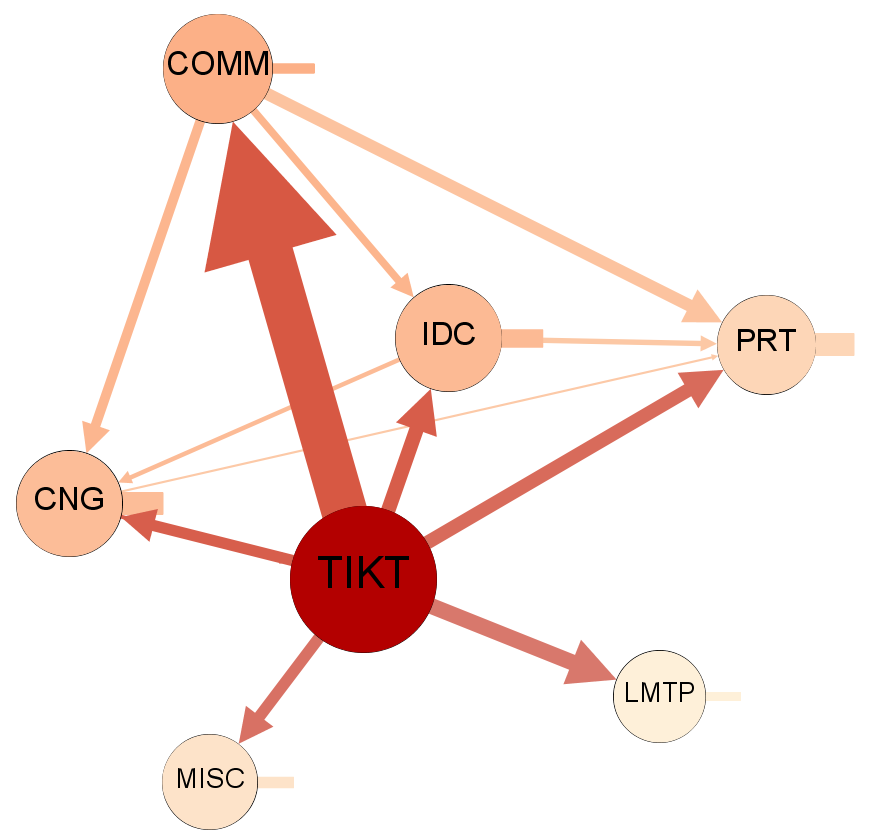}}%\vspace{-10pt}
	\caption{Visualization of infectivity matrix for ATM events. Each node denotes different event type. The edge direction indicates the influence direction and its width denotes the effect, the wider the stronger.}
	\label{fig:atm_infect}
\end{figure}

Table \ref{tab:performance} shows the averaged performance  of the proposed method compared to the alternatives. The confusion matrix for the seven event types are shown in Fig.~\ref{fig:confusion} by all methods. Not surprisingly, for both event type and timestamp prediction, our main approach, \emph{i.e.}, \textbf{ATRPP} outperforms by a notable margin. \textbf{ATRPP} report 0.634 F1 score and 3.92 MAE while \textbf{AERPP} reaches 0.617 F1 score and 3.98 MAE. Obviously, this verifies that synergically modeling event sequence and time series can boost the performance of predicting future events and time.Interestingly, all point process based models obtain better results on this task which suggests they are more promising compared to classical classification models. Indeed, our methodology provides an end-to-end learning mechanism without any pre-assumption in modeling point process. All these empirical results on real-world tasks suggest the efficacy of our approach, especially in capturing the temporal dynamics of events data.

\textbf{Visualization of influence pattern}. We visualize the infectivity matrix of \textbf{ATRPP} as in Fig.~\ref{fig:atm_infect}. Each node denotes one dimension which here represents one type of events. The directed edge means the influence strength from source node to destination node.  The size of nodes and depth of color is proportional to the weighted degree of nodes, which indicate the total influence of the node has on others. The width of edges is is proportional to the strength of influence. Self-loop edges are located at the right of nodes without an arrow. Note this setting applies to the subsequent two experiments. As is shown, it's obvious that TIKT (maintenance) have a strong influence over all types of errors (breakdown) as maintenance can greatly decrease the probability of breakdown of machines. Also, self-loop edge of TIKT node is too small to see which indicates that maintenance has low correlation itself. All types of errors have self-loop, indicating a recurrent pattern of errors. The breakdown of communication module (COMM) often leads to disfunction of cash dispenser module (CNG), printer (PRT) and internet data center (IDC). Besides, internet data center (IDC) problems influence cash dispenser module (CNG) and printer (PRT) weakly.

\subsection{Social network analysis}
\begin{figure}[htb!]
\centering
  \subfigure[\scriptsize MemeTracker]{\label{fig:meme_time}
  \includegraphics[width=0.23\textwidth]{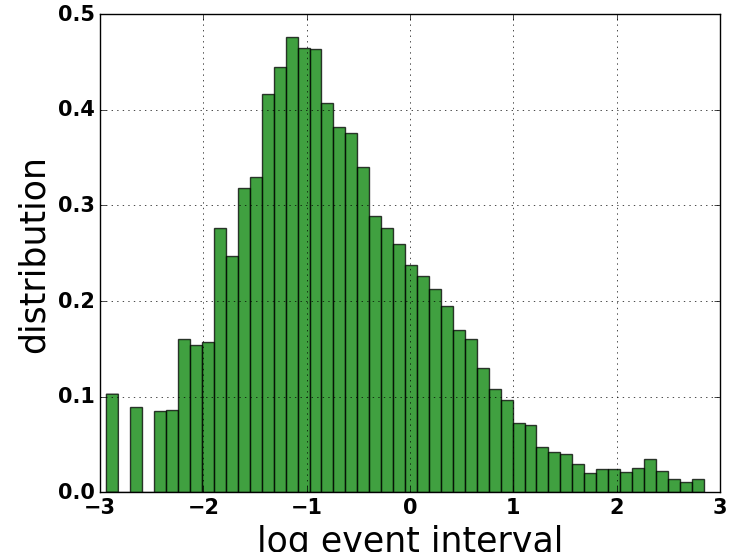}}
  \subfigure[\scriptsize MIMIC]{\label{fig:mimic_time}
  \includegraphics[width=0.23\textwidth]{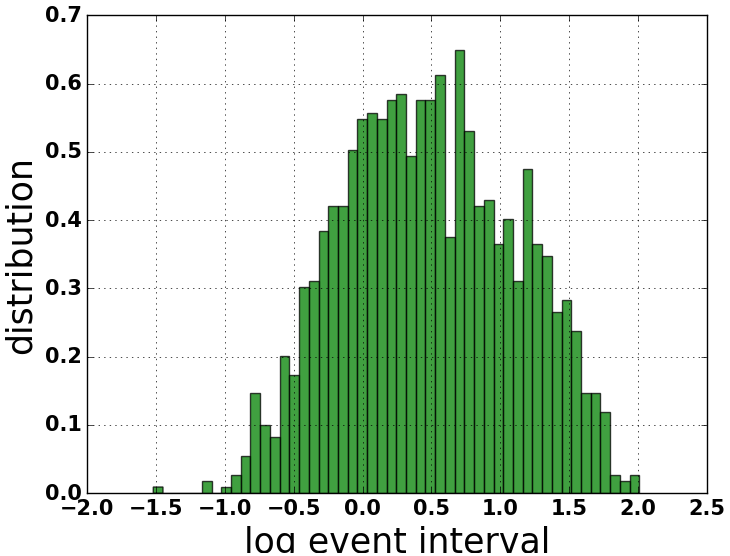}}
	\caption{Distribution of time interval between two events.}
	\label{fig:time_dis}
\end{figure}

In line with the previous works for information diffusion tracking~\cite{farajtabar2016multistage,ZhouAISTATS13,rodriguez2014uncovering}, the public dataset MemeTracker\footnote{http://memetracker.org} is used in this paper, which contains more than 172 million news articles or blog posts from various online media. The information, such as ideas, products and user behaviors, propagates over sites in a subtle way. For example, when Mark Zuckerberg posted "I just killed a pig and a goat", the meme appeared on the \emph{theguardian}, \emph{Fortune}, \emph{Business Insider} one after the other and it became viral. This cascade can be regarded as a realization of an information diffusion over the network. By looking at the observed diffusion history, we want to know through which site and when a particular meme is likely to spread. Besides, we want to uncover the hidden diffusion structure from those meme cascades, which is useful in other social network analysis applications, \emph{e.g.}, marketing by information maximization \cite{ZhuangICDM13}. From online articles, we collect more than 1 million meme cascades over different websites. For each meme cascade, we have the timestamp when sites mention a specific meme. For the experiments here, we use the top 500 media sites with the largest number of documents and select the meme cascades diffuse over them as done in previous works~\cite{farajtabar2016multistage,ZhouAISTATS13}. As a result, we obtain around 31 million meme cascades, which are randomly split into training (50\%), validating (\%20) and testing (\%30) parts. The event sequences are the meme cascades, which contain the website (dimension) and the timestamp. We count the times that a meme is mentioned during an hour over all the websites and use it as the time-series to reflect the hotness of the meme and the activity of websites.. The time interval of meme is shown in Fig.~\ref{fig:meme_time}.

As the ground truth of network is unknown, we proceed by following the protocol as designed and adopted in ~\cite{gomez2010inferring,gomez2011uncovering,ZhouAISTATS13}. We create a graph $G$, for which each node is a website. If a post on site $u$ has a hyperlink pointed to site $v$, then we create a directed edge $(u,v)$ with weight 1. If multiple hyperlinks exist between two sites, then the weight is accumulated. We use the graph as the ground truth and compare it with the inferred infectivity matrix from meme cascades.
The prediction performance is evaluated by \emph{Accuracy@k}, which evaluates whether the true label is within the top $k$ predicted dimensions.

\begin{table}[tb!]
\centering
\caption{Prediction evaluation by accuracy and MAE (mean absolute error) on MemeTracker dataset.}
\resizebox{0.48\textwidth}{!}{
\begin{tabular}{lrrr}
 \toprule
  model &accuracy@10 &accuracy@5 & MAE\\
 \midrule
  Poisson & ----- & ----- & 1.63\\
  SelfCorrecting & ----- & ----- & 1.70 \\
  Markov Chain & 0.563 &0.472 & ----- \\
  CTMC & 0.513 & 0.453& 1.69 \\
  Logistic & 0.463 &0.416& 1.72 \\
  Hawkes & 0.623 &0.563& 1.68 \\
  RMTPP & 0.679 & 0.589&1.55 \\
  TRPP & 0.681 & 0.592&1.52 \\
  ERPP & 0.673 & 0.586&1.56\\
  ATRPP & $\mathbf{0.694}$ & $\mathbf{0.598}$& $\mathbf{1.43}$\\
  AERPP & 0.678 & 0.589&1.45 \\
 \bottomrule
 \end{tabular}}
\label{tab:meme_performance}
\end{table}
\begin{figure}[htb!]
\centering
  \label{fig:meme_cluster4}
  \includegraphics[width=0.49\textwidth]{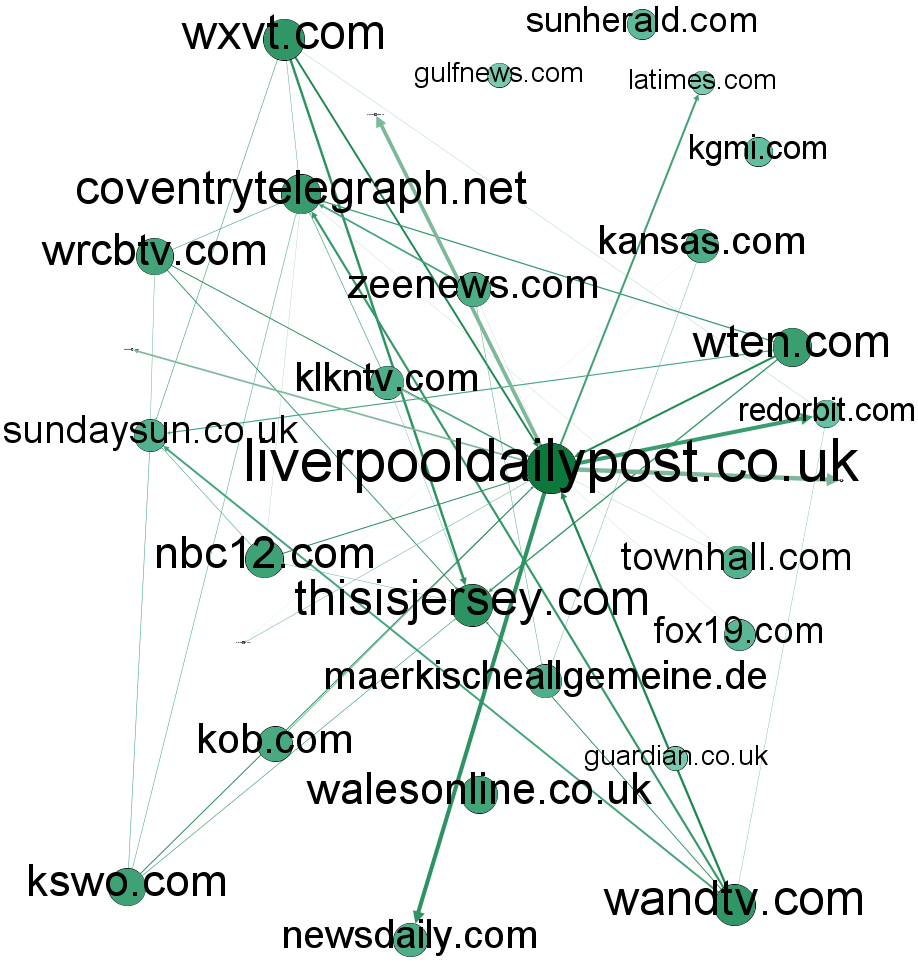}\\%\hspace{-2pt}
  \label{fig:meme_cluster11}
  \includegraphics[width=0.49\textwidth]{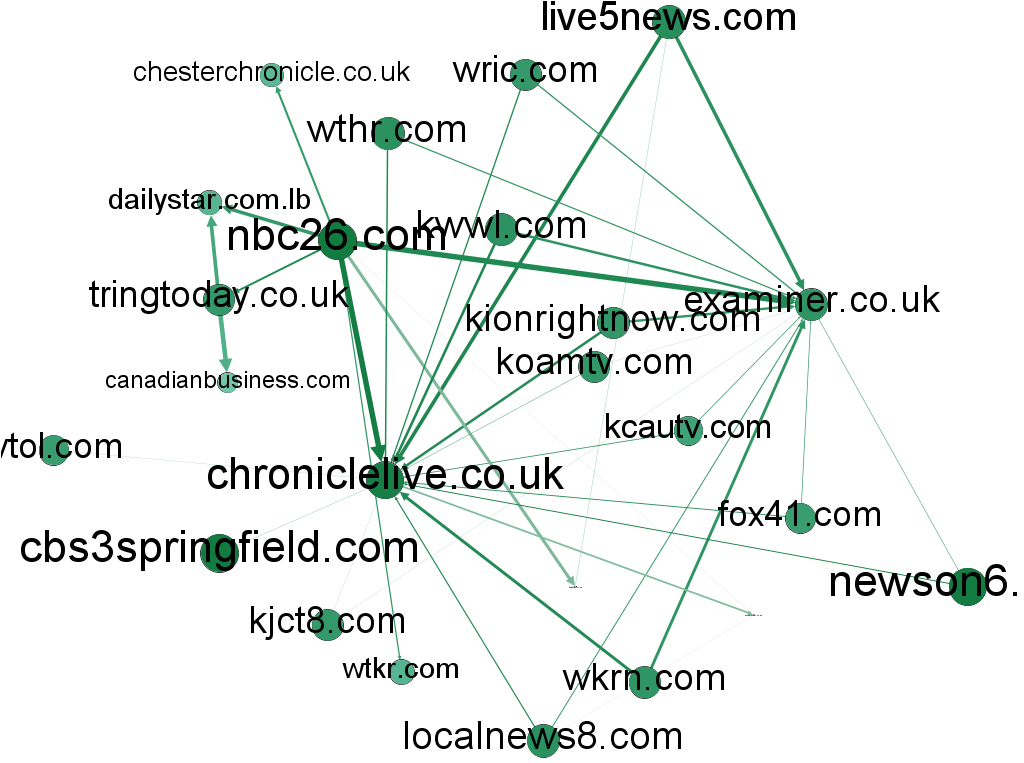}%\hspace{-2pt}
	\caption{Examples of detected media communities over the inferred diffusion network by our method ATRPP.}
	\label{fig:meme_graph}
\end{figure}
\begin{figure}[tb!]
	\centering
	\subfigure{\includegraphics[width=0.4\textwidth]{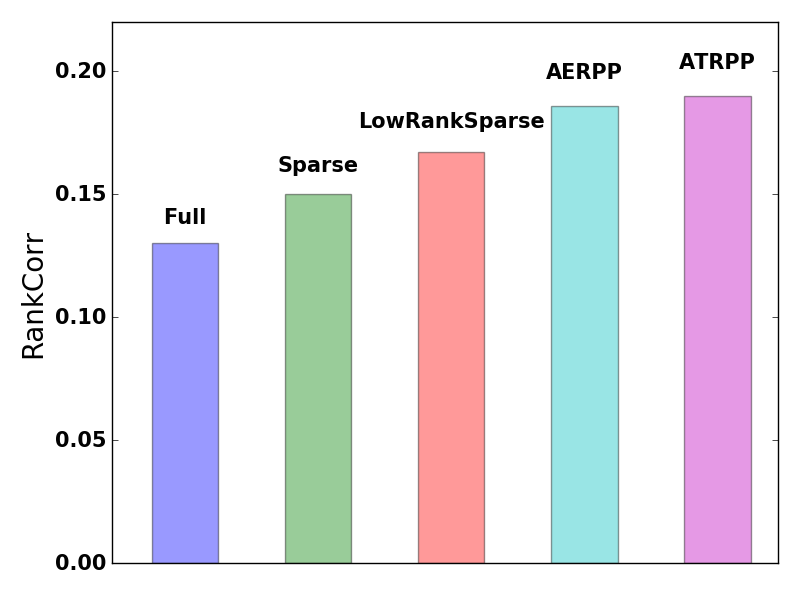}}%\vspace{-10pt}
	\caption{Performance measured by RankCorr on the MemeTracker dataset.}
	\label{fig:meme_rank}
\end{figure}

The prediction performance is shown in Table~\ref{tab:meme_performance} from which one can observe our model outperforms the alternatives. The rank correlation is shown in Fig.~\ref{fig:meme_rank}. Our models \textbf{ATRPP}, \textbf{AERPP} can better uncover the infectivity matrix than the competitive methods in terms of the correlation rank metric.

In order to visualize the learned network, we use community detection algorithm~\cite{blondel2008fast} with resolution 0.9~\cite{lambiotte2008laplacian} over learned directed network of \textbf{ATRPP}, which renders 18 communities. The resolution parameter controls the resolution of detected communities. It is set to lower values to get more communities (smaller ones), and is set higher to get fewer communities (bigger ones). Therefore, the communities can vary from the macroscale in which all nodes belong to the same community, to the microscale in which every node forms its own community. Fig.~\ref{fig:meme_graph} shows some examples of those communities. Some media domains, like \emph{liverpooldailypost.com} dominate in the cluster and what they publish usually spread to others and get viral.

\subsection{Electronic health records mining}
Uncovering the disease progression relation is important in healthcare analytics, which helps take preventive measures before fatal diseases happen. MIMIC-III (Medical Information Mart for Intensive Care III) is a large, publicly available dataset\footnote{https://mimic.physionet.org}, which contains de-identified health-related data during 2001 to 2012 for more than 40,000 patients. It includes information such as demographics, vital sign measurements, diagnoses and procedures. For each visit, a patient receives multiple diagnoses, with one as the primary one. We filter out 937 patients and 75 diseases. The age, weight, heart rate and blood pressure are used as time series signals of patients. The distribution of time intervals between every two visits is shown in Fig.~\ref{fig:mimic_time}. We have used
the sequences of 600 patients to train, 100 to evaluate and the rest for test.

\begin{table}[tb!]
\centering
\caption{Prediction evaluation on the MIMIC dataset.}
\resizebox{0.35\textwidth}{!}{
\begin{tabular}{lrr}
 \toprule
  model &accuracy & MAE\\
 \midrule
  Poisson & ------ & 0.562\\
  SelfCorrecting & ------ & 0.579 \\
  Markov Chain & 77.53\% & ------ \\
  CTMC & 73.62\% & 0.583 \\
  Logistic & 69.36\% & 0.643 \\
  Hawkes & 78.37\% & 0.517 \\
  RMTPP & 82.52\% & 0.546 \\
  TRPP & 82.26\% & 0.513 \\
  ERPP & 78.23\% & 0.521\\
  ATRPP & $\mathbf{85.23\%}$ & $\mathbf{0.497}$\\
  AERPP & 83.96\% & 0.503 \\
 \bottomrule
 \end{tabular}}
\label{tab:mimic_performance}
\end{table}
\begin{figure*}[htb!]
\centering
  \subfigure[\scriptsize Liver Diseases]{\label{fig:mimic_cluster8}
  \includegraphics[width=0.32\textwidth]{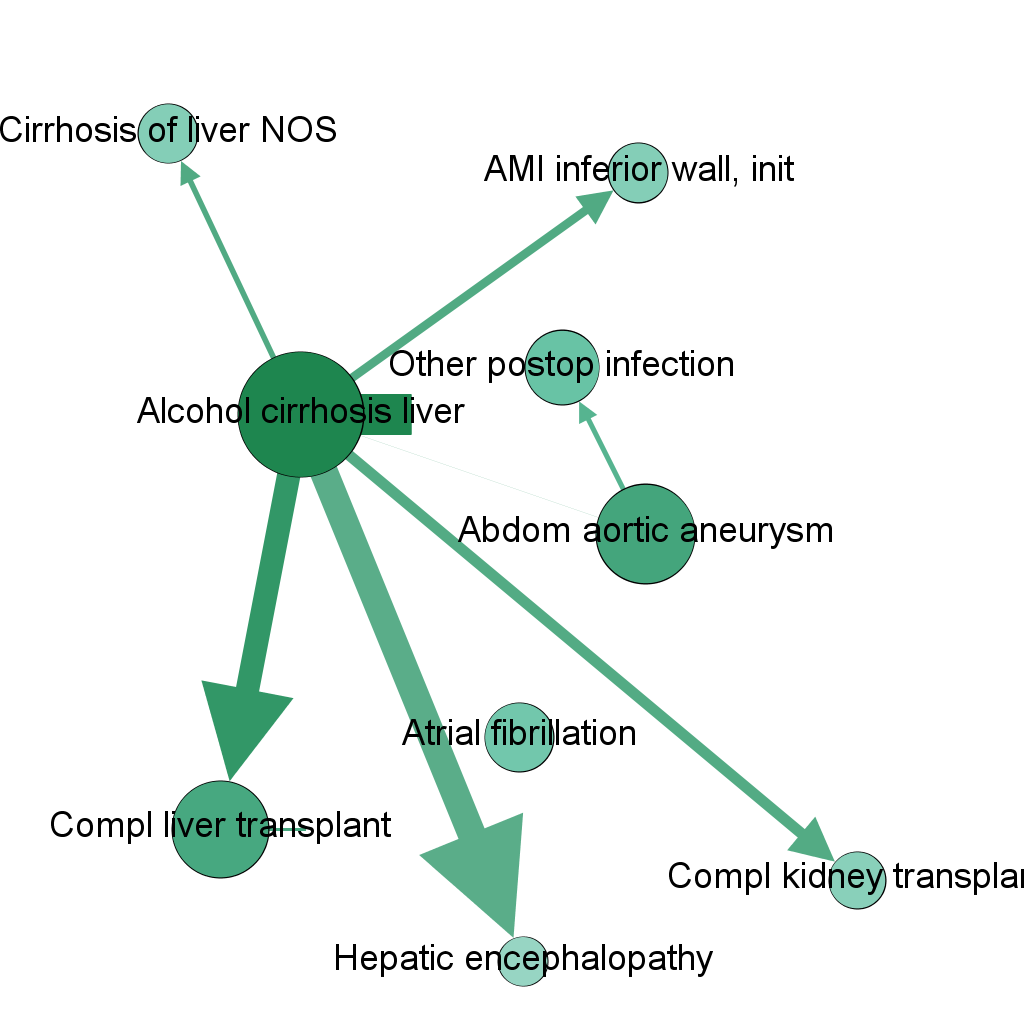}}%\hspace{-2pt}
  \subfigure[\scriptsize Respiratory diseases]{\label{fig:mimic_cluster3}
  \includegraphics[width=0.32\textwidth]{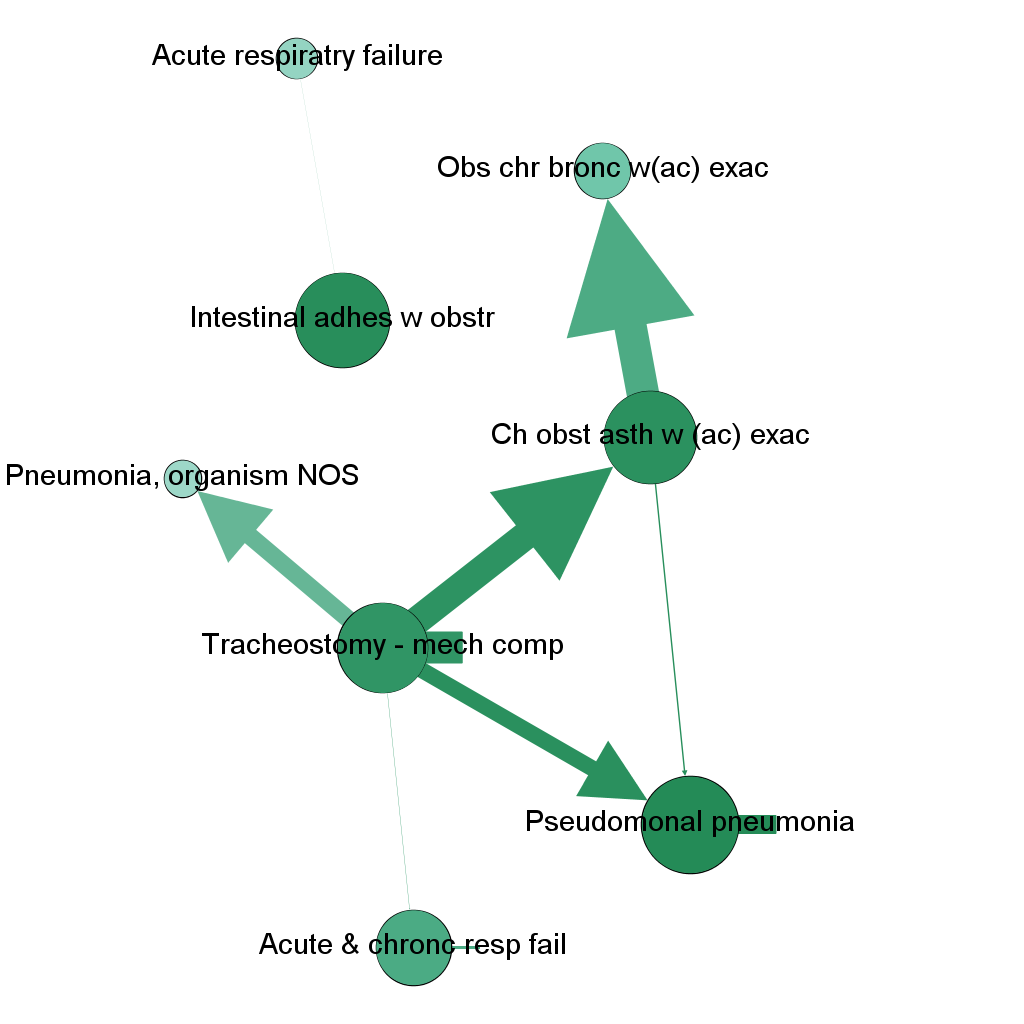}}%\hspace{-2pt}
  \subfigure[\scriptsize Alcohol-related diseases]{\label{fig:mimic_cluster6}
  \includegraphics[width=0.32\textwidth]{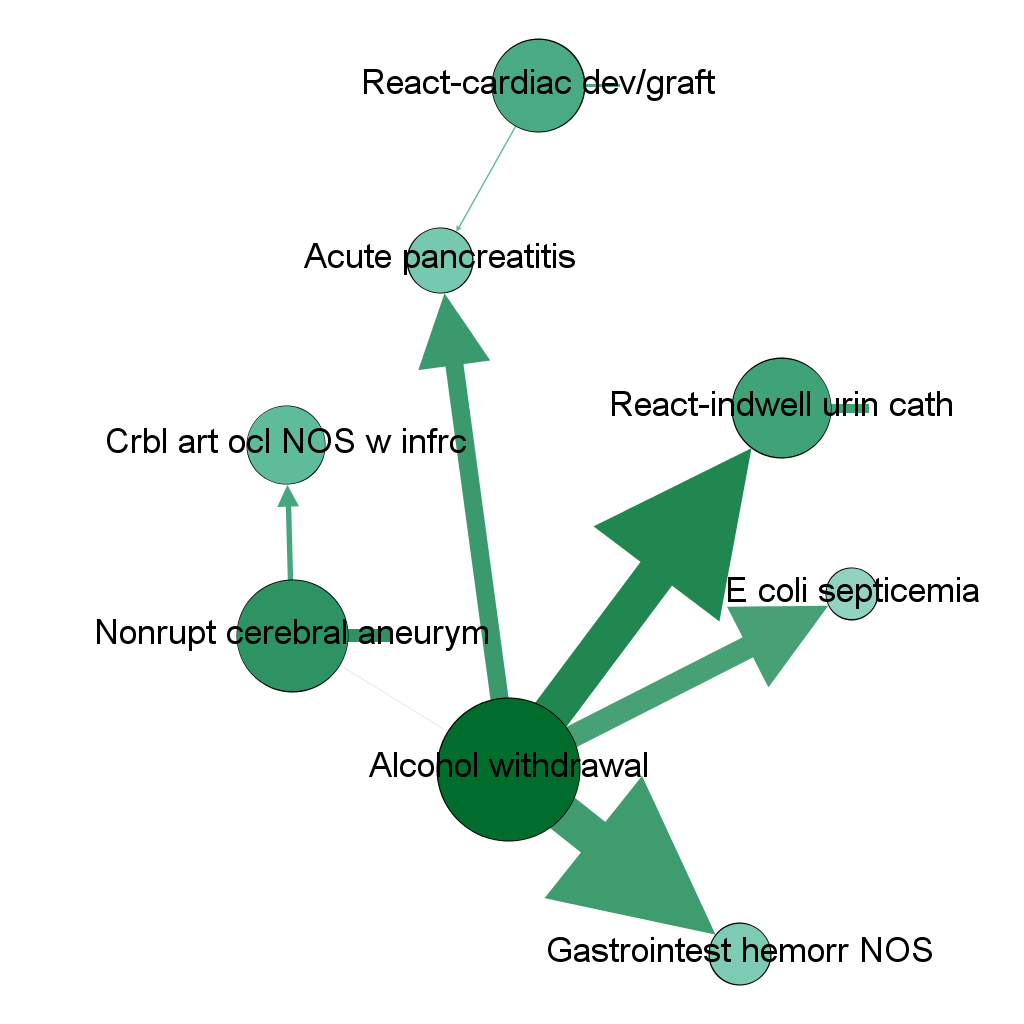}}\\%\vspace{-8pt}
  \subfigure[\scriptsize Heart and blood related diseases]{\label{fig:mimic_cluster2}
  \includegraphics[width=0.32\textwidth]{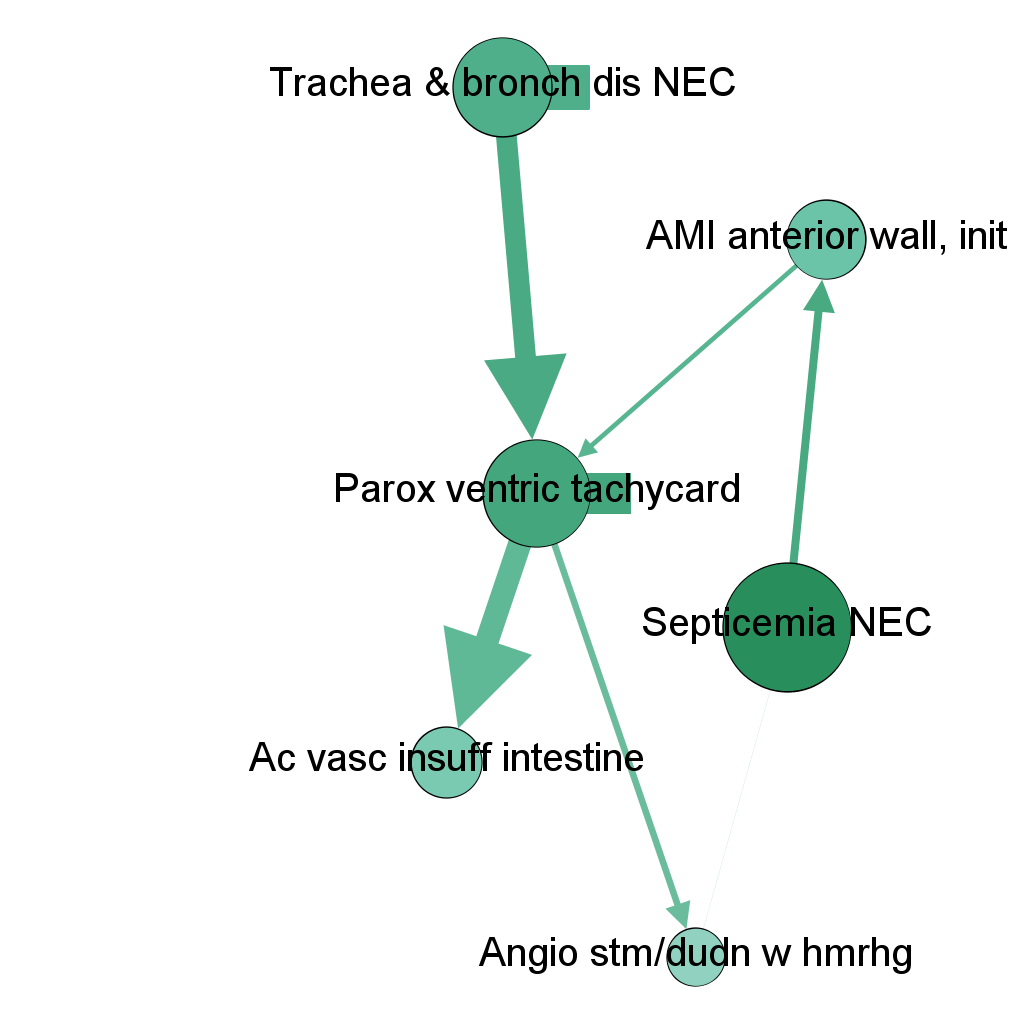}}%\hspace{-2pt}
  \subfigure[\scriptsize Blood metabolic diseases]{\label{fig:mimic_cluster7}
  \includegraphics[width=0.32\textwidth]{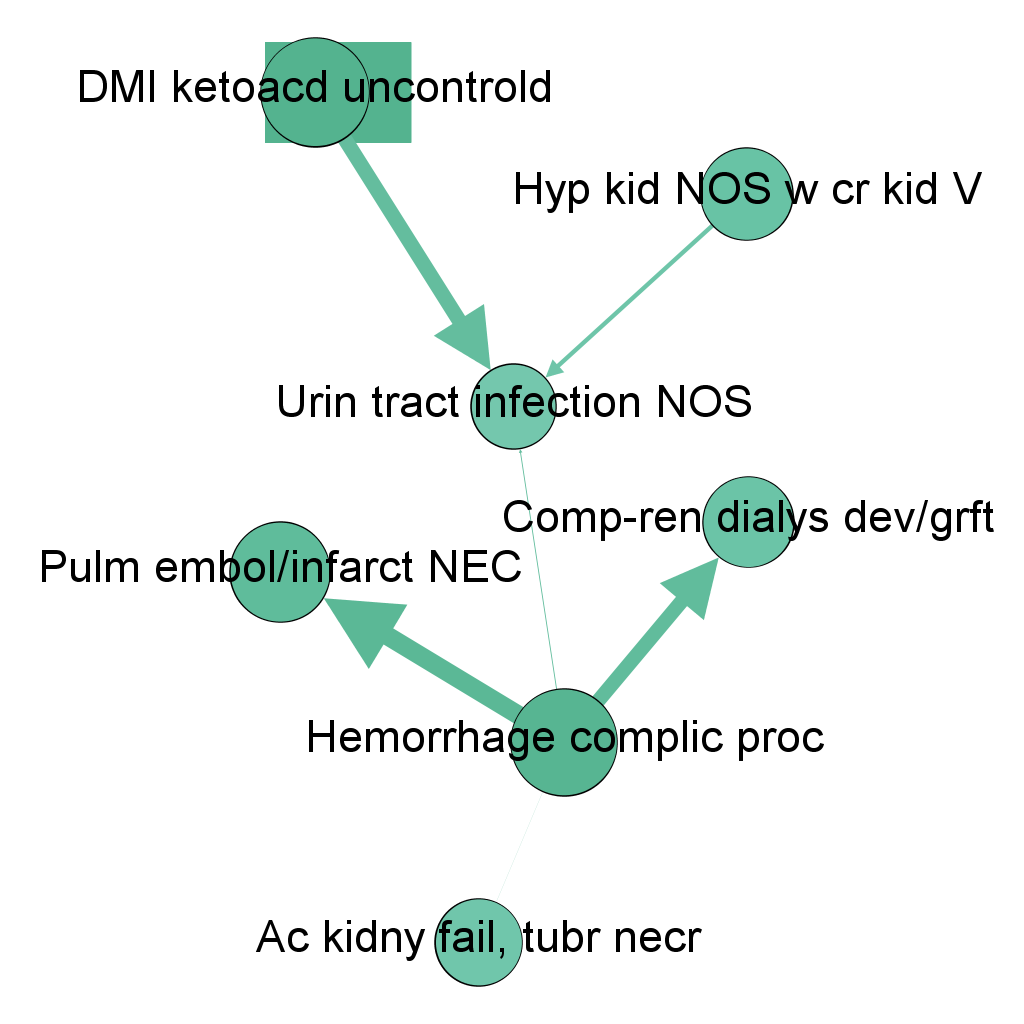}}%\hspace{-2pt}
\caption{Communities detected from the learned directed diseases network (these results have been qualitatively checked by the clinical experts based on their knowledge and exeprience). In Fig.~\ref{fig:mimic_cluster8}, AMI: Acute myocardial infarction of other anterior wall; postop: postoperative; Compl: Complications. In Fig.~\ref{fig:mimic_cluster3}, Ch obst asth w (ac) exac: Chronic obstructive asthma with (acute) exacerbation; bronc: bronchitis; adhes: adhesions; mech comp:mechanical complication; chronc resp fail: chronic respiratory failure. In Fig.~\ref{fig:mimic_cluster6}, React-indwell urin cath: reaction to indwelling urinary catheter; React-cardiac dev/graft: reaction to cardiac device and graft; E coli septicemia: Septicemia due to escherichia coli; hemorr: Hemorrhage; Crbl art ocl: Cerebral artery occlusion. In Fig.~\ref{fig:mimic_cluster2}, parox ventric tachycard: paroxysmal ventricular tachycardia; Ac vasc insuff intestine: Acute vascular insufficiency of intestine; Trachea \& bronch dis NEC: Other diseases of trachea and bronchus. In Fig.~\ref{fig:mimic_cluster7}, DMI: Diabetes with ketoacidosis type I; Hyp kid NOS w cr: hypertensive chronic kidney disease; dialys dev/grft: renal dialysis device \& graft; Pulm embol/infarct: pulmonary embolism and infarction.}
	\label{fig:mimic_graph}
\end{figure*}

Similar to MemeTracker, community detection algorithm~\cite{blondel2008fast} with resolution 0.9 is applied on the learned directed network from \textbf{ATRPP}. The results show cohesion within communities, which demonstrates the effectiveness of our attention mechanism. Note that some edges are too small to be visible. Specifically, Fig.~\ref{fig:mimic_cluster8} is about liver diseases. The node of \emph{alcohol cirrhosis} has a large self-loop edge, which means this disease generally repeats many times. Besides, it has a large edge towards \emph{complications of transplanted kidney} and \emph{hepatic encephalopathy}, which means \emph{alcohol cirrhosis} has a high probability of developing into the these two diseases. Fig.~\ref{fig:mimic_cluster3} is about respiratory diseases. Similarly, \emph{mechanical complication of tracheostomy} and \emph{pseudomonal pneumonia} have large self-loop edges, indicating they often relapse. \emph{Mechanical complication of tracheostomy} has an edge towards \emph{pseudomonal pneumonia} while the reverse, interestingly, does not exist. Fig.~\ref{fig:mimic_cluster6} shows the graph for alcohol-related diseases. Obsessed in alcohol has impact on \emph{reaction to indwelling urinary catheter} regarding urinary system and \emph{hemorrhage of gastrointestinal tract} regarding gastrointestinal system. Fig.~\ref{fig:mimic_cluster2} is about heart and blood related diseases. Three different parts of body form a progression line, consisting of \emph{diseases of trachea and bronchus} (respiratory disease), \emph{paroxysmal ventricular tachycardia} (heart rate disturbance) and \emph{acute vascular insufficiency of intestine} (intestinal disease). The other line is \emph{septicemia} leads to \emph{acute myocardial infarction of other anterior wall} (heart attack), which results in \emph{paroxysmal ventricular tachycardia} (heart rate disturbance). Fig.~\ref{fig:mimic_cluster7} is about blood metabolic diseases. \emph{Hemorrhage complicating a procedure} leads to \emph{pulmonary embolism and infarction} and \emph{complications due to renal dialysis device, implant, and graft}. \emph{Diabetes with ketoacidosis} results in \emph{urinary tract infection}. For the observed strong association as stated above, we conjecture it might be due to either causal or correlation relationship, which can provide supporting evidence and implication for clinical staff and is subject to further analysis by health practitioners.

Table \ref{tab:mimic_performance} reports the predictive performance of various models. \textbf{ATRPP} outperforms alternatives in both disease types and time prediction. Here first-order Markov Chain outperforms higher-order models, the reason might be due to the fact that visits of patients are sparse and there is not enough data to train the higher order Markov chains.
\begin{figure}[tb!]
\centering
  \includegraphics[width=0.49\textwidth]{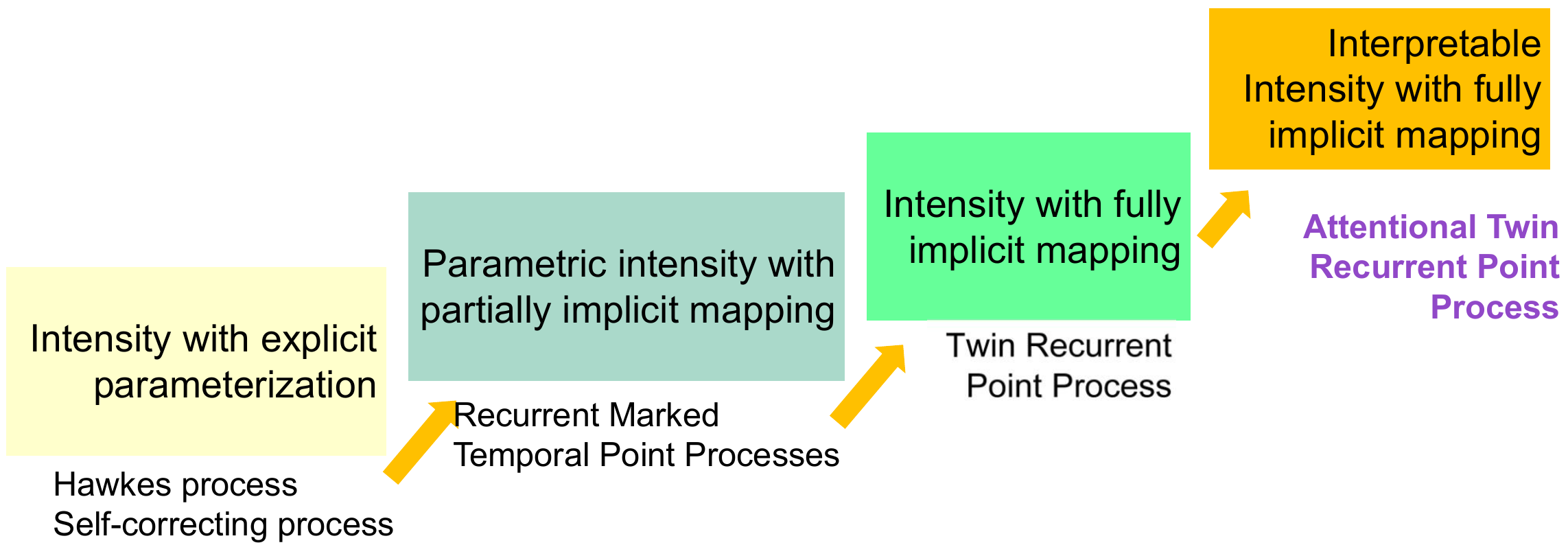}%\vspace{-10pt}
	\caption{The evolving of point process modeling. The embodiments of the first two blocks can be referred to \cite{ZhouAISTATS13} and \cite{DuKDD16} respectively. While the third and forth blocks are our conference version \cite{XiaoAAAI17} and this extended journal paper (from left to right).}
	\label{fig:evolve}
\end{figure}

\section{Conclusion}\label{sec:conclusion}
We conclude this paper with Fig.~\ref{fig:evolve} and identify our proposed method as a recurrent point process model. To elaborate, Hawkes process uses a full explicit parametric model and RMTPP misses the dense time series features to model time-varying base intensity, assumes a partially parametric form for it and model the pseudo multi-dimensional point process.
We make a further step by proposing an interpretable model which is simple and general and can be trained end-to-end. Most importantly, our model can uncover the subtle network structure and provide interpretable evidence for predicting result. The extensive experiments in this paper have clearly suggested its superior performance in synthetic and real-world data, even when we have no domain knowledge on the problem at hand. This is in contrast to existing point process models where an assumption about the dynamics is often needed to be specified beforehand.

%\appendices
\ifCLASSOPTIONcompsoc
  %\section*{Acknowledgments}
   %This research was partially supported by The National Key Research and Development Program of China (2016YFB1001003), NSFC (61602176, 61672231, 61527804, 61521062), STCSM (15JC1401700, 14XD1402100), China Postdoctoral Science Foundation Funded Project (2016M590337), the 111 Program (B07022) and NSF (IIS-1639792, DMS-1620345), NSFC-Zhejiang Joint Fund for the Integration of Industrialization and Informatization U1609220.
   \section*{Acknowledgments}
   The authors thank Robert Chen with Emory University School of Medicine, and  for helpful discussions and suggestions on the study of the computational experimental results on the MIMIC dataset. We are also thankful to Changsheng Li who shared us the ATM log data from IBM to allow us to perform the predictive maintenance study on real-world data.
\else
  %\section*{Acknowledgment}

\fi
\ifCLASSOPTIONcaptionsoff
  \newpage
\fi

\bibliographystyle{IEEEtran}
\bibliography{TKDE16Ref}
\end{document}